\documentclass[acmsmall]{acmart}

\AtBeginDocument{%
  }

\setcopyright{acmcopyright}
\copyrightyear{2018}
\acmYear{2018}
\acmDOI{XXXXXXX.XXXXXXX}

\acmJournal{JACM}
\acmVolume{37}
\acmNumber{4}
\acmArticle{111}
\acmMonth{8}

\usepackage{graphicx}
\usepackage{subfigure} 
\usepackage{amsmath}
\DeclareMathOperator*{\argmax}{arg\,max}
\DeclareMathOperator*{\argmin}{arg\,min}

\usepackage{amsthm}
\usepackage{amssymb}
\usepackage{algorithm}
\usepackage{algorithmic}
\usepackage{url}
\usepackage{float}
\usepackage{multirow}
\usepackage{color}
\usepackage{soul}
\usepackage{enumitem}
\usepackage{ulem}
\usepackage{caption}
\usepackage{makecell}
\usepackage{nicefrac}

\usepackage{todonotes}

\usepackage{hyperref}
\hypersetup{hidelinks,
	colorlinks=true,
	allcolors=black,
	pdfstartview=Fit,
	breaklinks=true}

\usepackage{array}
\newcolumntype{L}[1]{>{\raggedright\let\newline\\\arraybackslash\hspace{0pt}}m{#1}}
\newcolumntype{C}[1]{>{\centering\let\newline  \\\arraybackslash\hspace{0pt}}m{#1}}
\newcolumntype{R}[1]{>{\raggedleft\let\newline \\\arraybackslash\hspace{0pt}}m{#1}}

\newtheorem{remark}{Remark}[section]
\newtheorem{problem}{Problem}
\newtheorem{definition}{Definition}
\newtheorem{example}{Example}

\newcommand{\E}{\mathbb{E}}
\newcommand{\Hs}{\mathcal{H}}

\newcommand{\errapp}{\mathcal{E}_{\text{app}}}
\newcommand{\errest}{\mathcal{E}_{\text{est}}}
\newcommand{\erropt}{\mathcal{E}_{\text{opt}}}

\newcommand{\dtrain}{{D}_{\text{train}}}

\newcommand{\xii}{x_{i}}
\newcommand{\yii}{y_{i}}

\newcommand{\vect}[1]{\mathbf{#1}}

\makeatletter
\newcommand{\subalign}[1]{%
	\vcenter{%
		\Let@ \restore@math@cr \default@tag
		\baselineskip\fontdimen10 \scriptfont\tw@
		\advance\baselineskip\fontdimen12 \scriptfont\tw@
		\lineskip\thr@@\fontdimen8 \scriptfont\thr@@
		\lineskiplimit\lineskip
		\ialign{\hfil$\m@th\scriptstyle##$&$\m@th\scriptstyle{}##$\crcr
			#1\crcr
		}%
	}
}
\makeatother

\begin{document}

\title{Automated Machine Learning: From Principles to Practices}

%
\author{Zhenqian Shen}
\affiliation{%
 \institution{Tsinghua University}
 \streetaddress{1 Th{\o}rv{\"a}ld Circle}
 \city{Hekla}
 \country{China}}
\email{szg22@mails.tsinghua.edu.cn}

\author{Yongqi Zhang}
\affiliation{%
\institution{4Paradigm Inc.}
\country{China}}
\email{yzhangee@connect.ust.hk}

\author{Lanning Wei}
\affiliation{%
 \institution{Institute of Computing Technology, Chinese Academy of Sciences}
 \country{China}}
\email{weilanning1997@gmail.com}


\author{Huan Zhao}
\affiliation{%
 \institution{4Paradigm Inc}
 \country{China}}
\email{hzhaoaf@connect.ust.hk}

\author{Quanming Yao}
\authornote{Corresponding Author.}
\affiliation{%
 \institution{Tsinghua University}
 \city{Beijing}
 \country{China}
}
\email{qyaoaa@mail.tsinghua.edu.cn}


\renewcommand{\shortauthors}{Zhenqian Shen, Yongqi Zhang, Lanning Wei, Huan Zhao, and Quanming Yao}

\begin{abstract}

Machine learning (ML) methods have been developing rapidly, but configuring and selecting proper methods to achieve a desired performance is increasingly difficult and tedious. To address this challenge, automated machine learning (AutoML) has emerged, which aims to generate satisfactory ML configurations for given tasks in a data-driven way. In this paper, we provide a comprehensive survey on this topic. 
We begin with the formal definition of AutoML and then introduce its principles, including the bi-level learning objective, the learning strategy, and the theoretical interpretation. 
Then, we summarize the AutoML practices
by setting up the taxonomy of existing works based on three main factors: the search space, the search algorithm, and the evaluation strategy.
Each category is also explained with  the representative methods. 
Then, we illustrate the principles and practices with exemplary applications
from  configuring ML pipeline,
one-shot neural architecture search,
and integration with foundation models. 
Finally, we highlight the emerging directions of AutoML and conclude the survey.
\end{abstract}



\received{20 February 2007}
\received[revised]{12 March 2009}
\received[accepted]{5 June 2009}

\maketitle

\section{Introduction}
\label{sec:intro}

The past decades have witnessed the rapid development of machine learning techniques, evolving from Support Vector Machine~(SVM)~\cite{cortes1995support, ben2001support} that is designed for handling basic classification task, to Convolutional Neural Network~(CNN)~\cite{krizhevsky2012imagenet, he2016deep} that can well handle image data in computer vision tasks, to Large Language Model~(LLM)~\cite{ouyang2022training, touvron2023llama} that possesses remarkable ability of  general-purpose language generation. 
This evolution reflects a profound development marked by increasing complexity and capability within the domain of machine learning. However, as these techniques are applied in practical settings, the need for meticulous calibration of learning configurations, including adjustments in hyper-parameters and model structures, becomes necessary to attain desirable machine learning performance~\cite{bergstra2011algorithms, feurer2015efficient}. Nevertheless, the intricacy associated with optimizing such configurations presents a formidable challenge, calling for  considerable human effort and domain expertise~\cite{zoph2017neural, liu2018darts}.



\begin{figure}[t]
	\centering
	\includegraphics[width=1.00\textwidth]{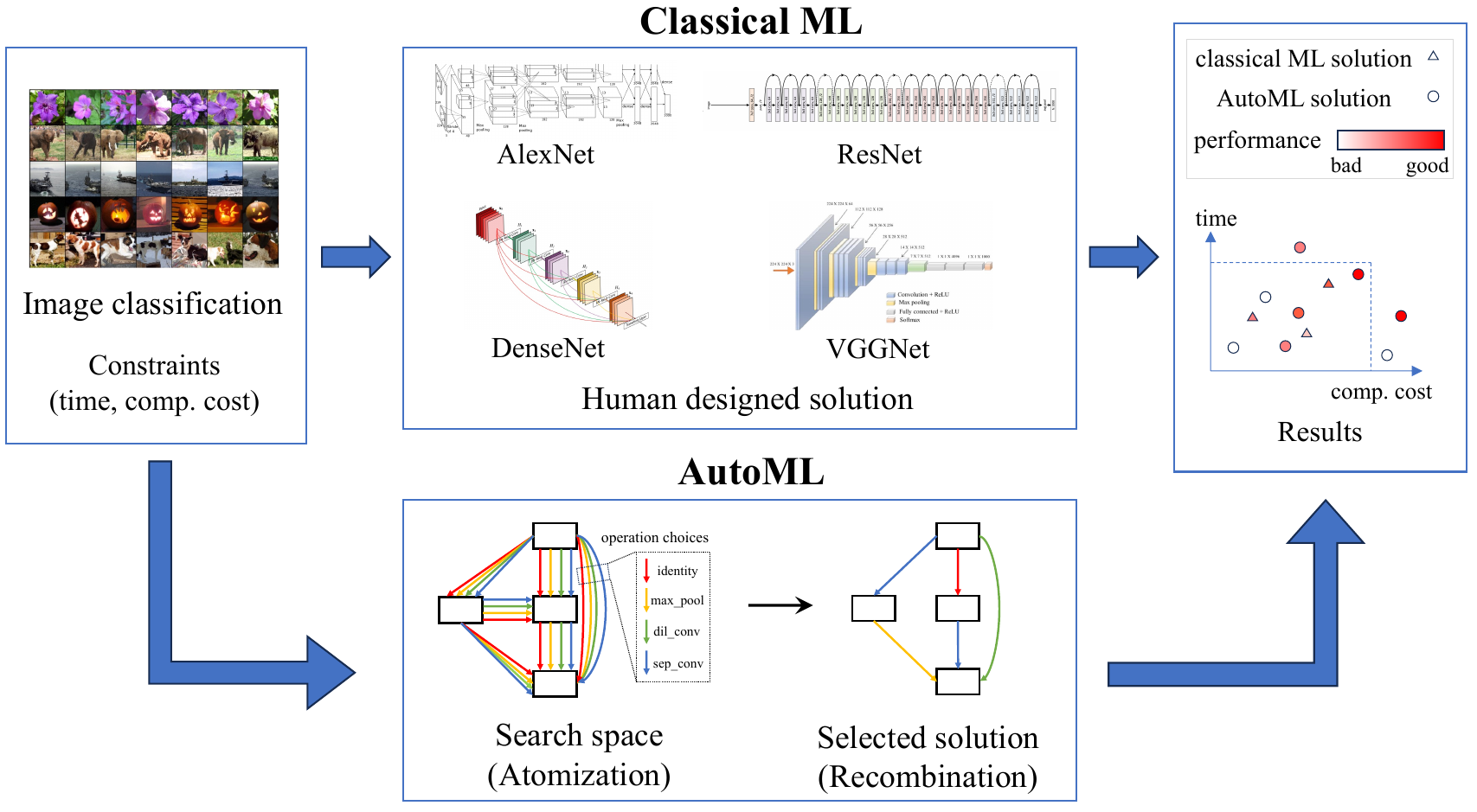}
	\caption{A brief illustration of AutoML framework~(take image classification as an example). For classical machine learning, solutions are designed by human experts to solve machine learning tasks. While for AutoML, it first atomizes the learning configurations and then recombines them to generate machine learning solutions. In this figure, time means inference time and comp. cost refer to computational cost.}
	\label{fig:frame}
\end{figure}

To tackle with this challenge, the concept of automated machine learning~(AutoML) has emerged as a feasible solution that attracts
both academic and industrial interest. 
Figure~\ref{fig:frame} illustrates the main idea of AutoML in machine learning problems.
For specific machine learning tasks with learning constraints~(i.e. inference time and computational cost limitation), classical machine learning methods utilize human designed solution. 
While AutoML first atomizes candidate solutions into a search space, and then 
considers these constraints and generates a satisfactory machine learning solution among the candidate solutions through recombination. 
Through that framework, AutoML significantly reduces the human labor cost and makes the machine learning process more accessible to a broader spectrum of users. 
Up to now, many AutoML methods has been proposed and they are applied in various real-world machine learning problems, such as image classification~\cite{zoph2017neural, liu2018darts} and machine translation~\cite{so2019evolved}.

In this survey, we give a comprehensive overview of AutoML, from principles to practices.
We first introduce the formal definition of AutoML and then discuss the principles of AutoML, including the bi-level learning objective, the learning strategy, and the theoretical interpretation. 
Several relevant topics of AutoML are also discussed. 
Then, we present the AutoML practices through the taxonomy based main factors of AutoML, where representative AutoML methods in each category are introduced. 
With the combination of AutoML principles and practices, we discuss exemplary applications and emerging directions of AutoML. 
The main contributions of the survey are summarized as follows:
\begin{itemize}[leftmargin=*]

\item We give a formal definition of AutoML,
which is inspired by the definition of classical machine learning~\cite{tom1997machine, mohri2018foundations, zhou2021machine}. 
The definition highlights learning configuration atomization and recombination steps in the AutoML process,
which enables us to derive principles and practice based on existing works.

\item We introduce the AutoML principles. We discuss the bi-level learning objective of AutoML
and the general learning strategy. 
We also give a theoretical interpretation of AutoML through error decomposition. The theoretical interpretation explains how AutoML can improve machine learning performance and the corresponding trade-offs. 
The principles of AutoML reveals the problem decomposition and solution generation strategy in AutoML. 
They can further provide guidance for the design of AutoML methods in practice.


\item We summarize AutoML practices through the taxonomy of main factors of AutoML, i.e., search space, search algorithm and evaluation strategy. 
We further provide the representative works in each category. 
The practices of AutoML demonstrate the current research progresses in AutoML and enhance better understanding of existing AutoML methods. 




\item We show light upon emerging directions of AutoML from the perspectives of problem setup,  techniques, theory and applications. 
These discussions reveal current weaknesses of AutoML and provide guidance for future AutoML research.
\end{itemize}

Compared to other surveys that are related to AutoML~\cite{elsken2019neural, yu2020hyper, he2021automl, white2023neural}, our survey focuses more on the principle analysis of AutoML. We give a formal definition of AutoML, formulate the AutoML problem as a bi-level optimization problem and present a theoretical analysis of AutoML. 
There are surveys that focus on specific topics of AutoML, e.g., Neural architecture search~(NAS)~\cite{elsken2019neural, white2023neural} that automates the design of neural network architecture and Hyper-parameter optimization~(HPO)~\cite{yu2020hyper} that tunes hyper-parameters for machine learning models to optimize its performance. 
It is worth noting that our survey does not exhaustively cover the entire spectrum of AutoML techniques.
Our primary objective is to summarize the key principles in AutoML and provide reference for real-world AutoML practices. 


\section{Principles}
\label{sec:overview}

In this section, we first define the AutoML problem in Section~\ref{ssec:prodef}.
In Section~\ref{sec:lob_stra}, we discuss the universal learning objective and general learning strategy of AutoML
In Section~\ref{sec:theory}, we provide the theoretical interpretation of AutoML.
Taxonomies of existing works according to search space, 
search algorithm and evaluator are presented in Section~\ref{sec:cate}.
Finally, we discuss the relevant topics of AutoML in Section~\ref{sec:rel_topic}.

\subsection{Definition: from ML to AutoML}
\label{ssec:prodef}

First,
we recall the formal definition of machine learning. 

\begin{definition}[Machine learning~\cite{tom1997machine, mohri2018foundations, zhou2021machine}] 
\label{def:machlea}
A computational method is said to  learn from experience $E$ with respect to some classes of task $T$ 
and performance measure $P$
if its performance can improve with $E$ on $T$ measured by $P$.
\end{definition}

For example, in an object detection task~($T$) targeting at locating instances of objects in images~\cite{ren2015faster, redmon2016you}, 
experience~($E$) corresponds to images with its labeled bounding box, and $P$ measures the performance of 
object detection, 
such as Average Precision~(AP) and mean Average Precision~(mAP)~\cite{girshick2015fast, ren2015faster}. 
Another example is molecular property prediction~($T$)~\cite{wu2018moleculenet, yang2019analyzing}, which targets at predict the property of molecules precisely. 
In this application, experience~($E$) corresponds to molecules with its property~(label)
and $P$ measures the performance of molecular property prediction, such as
accuracy or AUC~\cite{wu2018moleculenet}. 
Generally, the performance~($P$) will be improved with more experience~($E$). 
Although the machine learning is now widely utilized in many real world problems~\cite{krizhevsky2012imagenet, redmon2016you, kenton2019bert}, 
classical machine learning still faces unavoidable challenges 
as follows: 
\begin{itemize}[leftmargin=*]
\item 
Complex design choices: 
the design choices of machine learning methods include hyper-parameters, complex features and diverse architectures.
The human designed methods are usually only the scattered points in the potential design space, 
which are almost impossible to obtain the optimum. 

\item
High human labor cost: 
most top challenges for the adoption of machine learning is closely related to human labor, which can be physical cost or domain expertise to realize machine learning system. 
However, these kinds of human labor are extremely costly, which severely affects wide implementation of machine learning in real-world scenarios. 
\end{itemize}

\begin{table}[t]
	\footnotesize
	\caption{Comparison between classical machine learning~(ML) and AutoML through examples.}
	\vspace{-10px}
	\begin{tabular}{C{45px}|C{148px}|C{167px}}
		\toprule
		problems & Classical ML & AutoML \\ \midrule
		Configuring ML Pipelines   & Scikit-learn~\cite{scikit-learn}: a machine learning library, require human experts to select appropriate learning configurations for specific problems based on prior knowledge 
		& Auto-sklearn~\cite{feurer2015efficient}: a method that automatically search for proper data preprocessor, feature preprocessor and classifier for a machine learning problem and build ensembles among models to obtain the best performance \\ \midrule
		Network Architecture Design & AlexNet~\cite{krizhevsky2012imagenet}: human designed model architecture that only have a few learning configurations can be manually adjusted
		& EfficientNet~\cite{tan2019efficientnet}: 
		a family of network architectures designed to achieve better trade-offs between model size and performance by automatically scaling the network's depth, width, and resolution
		\\ \midrule
		Foundation Models     & 
		Adapter~\cite{houlsby2019parameter}: 
		a type of fine-tune strategy which needs human experts to adjust configurations when used in different pre-train models.
		& AutoPEFT~\cite{zhou2023autopeft}: a method which uses multi-dimensional Bayesian optimization to search for a set of suitable configurations of fine-tuning strategies for the corresponding pre-train model \\ \bottomrule
	\end{tabular}
	\label{tab:example}
\end{table}

To address above problems of classical machine learning, the idea of AutoML is proposed.
It provides machine learning methods with more flexible design choices and automatically finds 
the optimum. 
Formally, we define AutoML as follows:

\begin{definition}[AutoML] 
\label{def:automl}
A computational method attempts to atomize (all or a part of) learning configurations and 
then recombine them through optimizing
performance measure $P$ with experience $E$ on some classes of task $T$.
\end{definition}

According to the definition of AutoML, 
we can infer that AutoML 
targets at addressing complex design choices and high human labor cost in classical machine learning.
Through the process of atomization of learning configurations, 
AutoML offers a broader range of design choices in the design space compared to classical machine learning.
By recombining these learning configurations through optimization, 
AutoML largely reduces human labor in exploring domain-specific prior knowledge.  
To better understand the difference between classical machine learning and AutoML, 
here we give several application examples in 
Table~\ref{tab:example}. 

\subsection{Learning Objective and Strategy}
\label{sec:lob_stra}


As above, 
AutoML employs a two-step approach,
i.e., atomization and recombination,
to addresses the challenges of classical machine.
To facilitate this process, we introduce the learning configuration
parameter $\alpha $, which consists of atomized components of classical machine learning solutions.
Additionally, we define the configuration 
search space $\mathcal{A}$, 
which includes the value range $A$ for individual configurations.
By sampling a configuration $\alpha$ from the search space $A$, we can construct the recombined learning method $C_\alpha$.
Based on these settings, 
we introduce the bi-level learning objective of AutoML as follows:

\begin{figure}[t]
	\centering
	\includegraphics[width=0.90\textwidth]{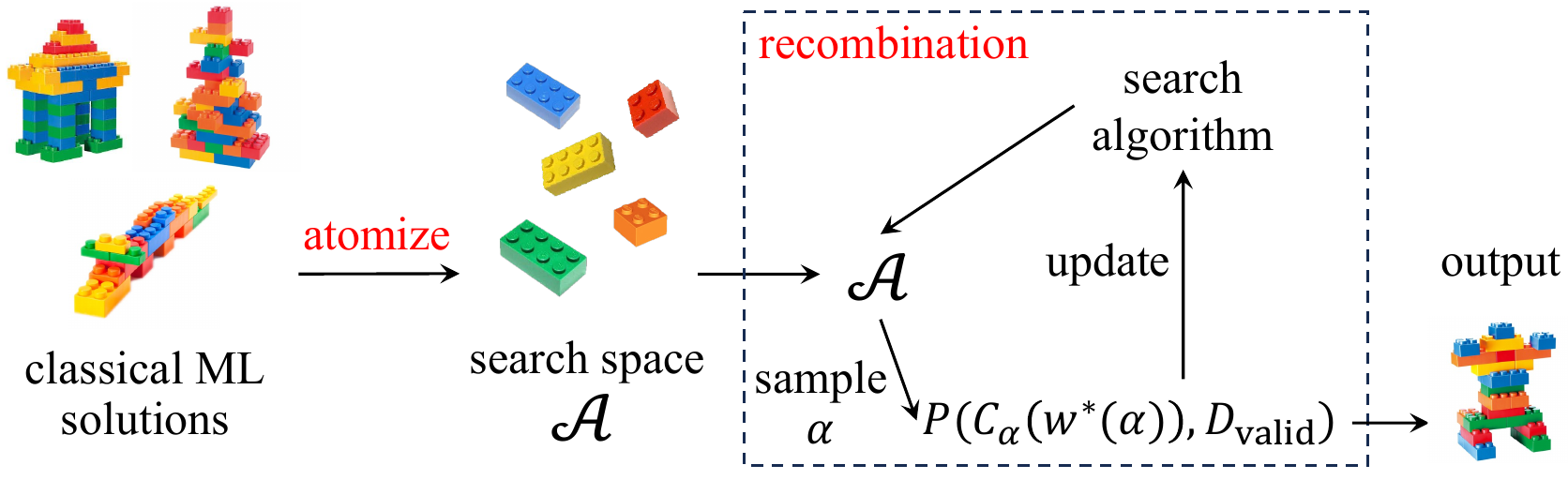}
	\caption{General pipeline to handle AutoML. Here LEGO blocks refer to atomized learning configurations. The shape formed by building blocks refer to the machine learning solutions. }
	\label{fig:rela_fact}
\end{figure}

\begin{problem}
\label{prob:bi_obj}
Bi-level learning objective of AutoML: 
\begin{align}
	\label{Eq:bi_automl1}
	\alpha^*
	= & \argmax\nolimits_{\alpha \in \mathcal{A}} P(C_{\alpha}(w^*(\alpha)), \mathcal{D}_{\text{valid}}),  
	\\
	\label{Eq:bi_automl2}
	\text{s.t.} &
	\begin{cases}
		w^*(\alpha) = \argmin\nolimits_{w} L(C_{\alpha}(w), \mathcal{D}_{\text{train}})
		\\ 
		\quad 
		G(\alpha) \leq 0
	\end{cases},
\end{align}
where $\mathcal{D}_{\text{train}}$ (resp. $\mathcal{D}_{\text{valid}}$) corresponds to training (resp. validation) dataset, 
$w$ is the model parameter, $P$ is validation evaluation metric and $L$ is a loss function.
Here, 
$G(\alpha)$ refers to the constraints of this problem, including computation time, memory budget, 
inference time, communication costs, supported system level operations. 
\end{problem}

Then,
we introduce the general learning strategy that solves objectives~\eqref{Eq:bi_automl1} and \eqref{Eq:bi_automl2}. 
The main strategy, focusing on the upper level objective,
is shown in 
Figure~\ref{fig:rela_fact}. 
AutoML first atomizes 
the classical machine learning solutions~(the constructed shapes in the figure) and
form the configurations choices~(the LEGO blocks in the figure) into search space $\mathcal{A}$. 
After atomization, AutoML conducts the recombination step: a specific configuration $\alpha$ is firstly sampled from the search space to obtain the learning method $C_\alpha$. 
Then the learning method is trained for the optimized parameter $w^*(\alpha)$ and evaluated on the validation dataset. 
Based on the performance of $C_\alpha(w^*(\alpha))$, the search algorithm is adapted to guide the choice of next configuration $\alpha$.

From that learning procedure, we can summarize the three main factors of AutoML: 
\textit{search space}, 
\textit{search algorithm} and 
\textit{evaluation strategy}. 
In the upper level objective of~\eqref{Eq:bi_automl1}, search space corresponds to 
range
of $\alpha \in \mathcal{A}$, 
search algorithm corresponds to optimization of $\alpha$ in the upper level objective, 
and evaluation corresponds to 
obtaining the evaluation metric $P(C_{\alpha}(w^*(\alpha)), \mathcal{D}_{\text{valid}})$. 
Designing a good AutoML system requires careful consideration of all these three factors, which set up three core issues of AutoML: 
\begin{itemize}[leftmargin=*]
\item how to design a good search space? 
\item how to design an effective search algorithm?
\item how to evaluate model performance efficiently? 
\end{itemize}
In Table~\ref{tab:expfactor} below, 
we illustrate the three main factors in based on examples in 
Table~\ref{tab:example}.


\begin{table}[ht]
	\footnotesize
	\caption{Three main factors in AutoML problem examples. For search algorithm and evaluation, we give some techniques that are commonly used. }
	\centering
	\vspace{-10px}
	\begin{tabular}{C{80px}|C{107px}|C{70px}|C{87px}}
		\toprule
		AutoML problems & search space & search algorithm & evaluation \\ \midrule
		Configuring ML Pipelines~\cite{bergstra2012random, bergstra2011algorithms, li2017hyperband, domhan2015speeding}    & Common hyper-parameter of ML pipeline, e.g. learning rate, batch size &
		E.g. random search~\cite{bergstra2012random}, Bayesian optimization~\cite{bergstra2011algorithms}&
		E.g. early stopping~\cite{li2017hyperband}, learning curve exploration~\cite{domhan2015speeding}
		\\ \midrule
		Network Architecture Design~\cite{zoph2017neural, liu2018darts, mellor2021neural} & Network operation choices and possible connections between operations& 
		E.g. reinforcement learning~\cite{zoph2017neural}, gradient descent~\cite{liu2018darts}& 
		E.g. weight sharing~\cite{liu2018darts}, surrogate model~\cite{mellor2021neural} \\ \midrule
		Foundation Models~\cite{wang2023search, zhou2023autopeft, arango2024quicktune, wang2023cost}         & Candidate pre-train and fine-tuning methods, parameters and hyper-parameters of foundation models& 
		E.g. gradient-based methods~\cite{wang2023search}, Bayesian optimization~\cite{zhou2023autopeft}& 
		E.g. performance predictor ~\cite{arango2024quicktune}, bandit-based method~\cite{wang2023cost}
		\\ \bottomrule
	\end{tabular}
	\label{tab:expfactor}
\end{table}

\subsection{Theoretical Interpretation}
\label{sec:theory}

In this part, we will analyze AutoML theoretically through error decomposition and investigate how AutoML improves machine learning performance. 
Denote the data as input-output pairs $(x,y)$ endowed with probability distribution $p(x,y)$. 
Given a learner $h$, 
we want to minimize its \emph{expected risk} $R$,
i.e.,
$
R(h) = \int \ell(h(x),y)\ dp(x,y)=\E[\ell(h(x),y)]
$,
which is the expected loss w.r.t data distribution $p(x,y)$.
As we can only have 
a finite sample size in practice,
$R(h)$ is usually approximated with \textit{empirical risk}, i.e.,
$R_I(h) = \nicefrac{1}{I} \sum\nolimits_{(\xii,\yii) \in \dtrain} \ell(h(\xii),\yii)$,
which is the average of sample losses over the training set $\dtrain$ with sample size $I$. 
As learning configurations are sampled and evaluated iteratively, we denote $\alpha_i$ as the learning configuration in the $i$-th lower level learning step, $\mathbf{H}_i$ as the corresponding hypothesis space and totally $n$ configurations are sampled.  
For illustration, define
$\hat{h}=\arg\min_{h} R(h)$
be the function that minimizes the expected risk;
$h^*=\arg\min_{h \in \mathbb{H}} R(h)$	be the function in hypothesis space $\mathbb{H}$ that minimizes the expected risk;
$h^n=\arg\min_{h \in \mathbf{H}_n} R(h)$
be the function in hypothesis space $\mathbf{H}_n$ that minimizes the expected risk;
$h_I^n=\arg\min_{h \in \mathbf{H}_n} R_I(h)$
be the function in hypothesis space $\mathbf{H}_n$ that minimizes the empirical risk.
The \emph{total error} in AutoML can be decomposed as: 
\begin{align}
\mathcal{E}_{\text{total}} = \mathbb E[R(h_I^n) - R(\hat{h})]
= \underbrace{\E[R(h^*) - R(\hat{h})]}_{\displaystyle\errapp(\mathbb{H})} 
+ \underbrace{\E[R(h^n)-R(h^*)]}_{\displaystyle\mathcal{E}_{\text{opt}}(\mathbf{H}^n, \epsilon^n )}
+ \underbrace{\E[R(h^n_I)-R(h^n)]}_{\displaystyle\errest(\mathbf{H}^n,I)}.
\end{align}


\begin{itemize}[leftmargin=*]
	\item The \emph{approximation error} $\errapp(\mathbb{H})$ measures how close the
	optimal function $h^* \in \mathbb{H}$ can approximate the optimal learner $\hat{h}$.
	\item The \emph{optimization error} $\mathcal{E}_{\text{opt}}(\mathbf{H}^n, \epsilon^n )$ measures the distance between the optimal function $h^n$ in $\mathbf{H}^n$ to the optimal function $h^*$ in $\mathbb{H}$.
	\item
	The \emph{estimation error} $\errest(\mathbf{H}^n,I)$ 
	measures the effect of minimizing the empirical risk $R'(h)$ instead of the expected risk $R(h)$
	within $\mathbf{H}$.
\end{itemize}

The total error can be affected by the three main factors of AutoML. 
The larger search space expands the hypothesis space $\Hs$, while the search algorithm and evaluation mainly influence the optimization procedure of the 
bi-level problem~\eqref{Eq:bi_automl1}-\eqref{Eq:bi_automl2}. 
Table~\ref{tab:factor_err} summarizes the evolution of decomposed errors and computation time with changes of three
factors. 
When the search space becomes larger, the approximation error will drop due to larger hypothesis space. 
The 
optimization error and computation time will increase because it is more difficult to find optimum in larger hypothesis space. 
More iterations for search algorithm will lower 
optimization error and increase computation time.
Similarly, more per-iteration cost evaluation strategy will lower optimization error and increase computation time. 

\begin{table}[ht]
	\caption{Typical variations when search space, search algorithm, evaluation strategy changes. }
	\vspace{-10px}
	\footnotesize
	\begin{tabular}{@{} c| C{50px} | C{75px} | C{100px} @{}}
		\toprule
		& search space (larger) & search algorithm  (more iterations) & evaluation strategy (more per-iteration cost) \\ \midrule
		approximation error            & $\searrow$            & -                                   & -                                             \\ \midrule
		optimization error & $\nearrow$            & $\searrow$                          &  $\searrow$                                            \\ \midrule
		estimation error   & -                     & -                                   & -                                             \\ \midrule
		computation time               & $\nearrow$            & $\nearrow$                          & $\nearrow$                                    \\ \bottomrule
	\end{tabular}
	\label{tab:factor_err}
\end{table}

\begin{figure}[t]
	\centering
	\includegraphics[width=0.8\textwidth]{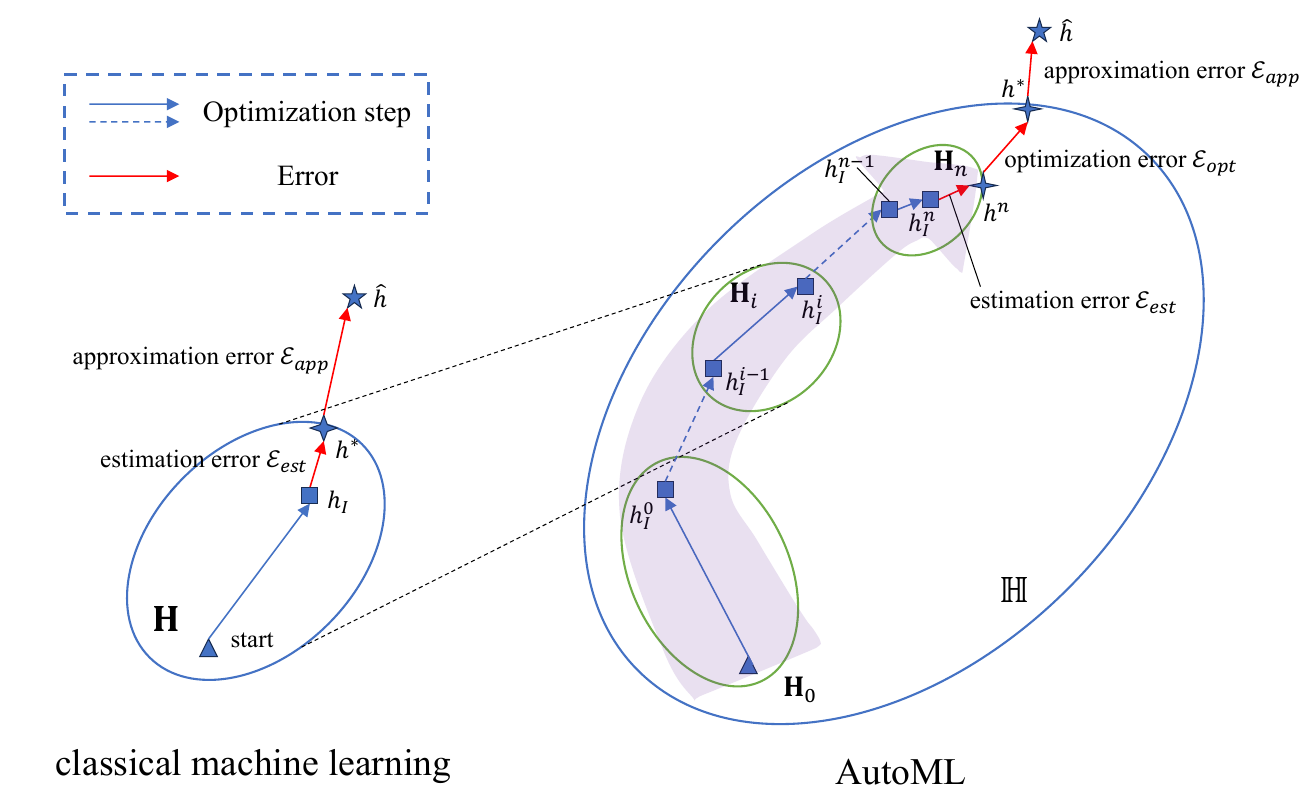}
	\caption{Comparison of error decomposition between classical machine learning and AutoML.}
	\label{fig:theory}
\end{figure}

\begin{remark}
Recall  in classical machine learning~\cite{bottou2008tradeoffs},
the \emph{total error} is as follows:
\begin{align}
	\mathcal{E}_{\text{total}} 
	= \underbrace{\E[R(h^*) - R(\hat{h})]}_{\displaystyle\errapp( \mathbf{H} )} 
	+ 
	\underbrace{\E[R(h_I)-R(h^*)]}_{\displaystyle\errest( \mathbf{H}, I )}.
\end{align}
We can find the error decomposition in classical machine learning is a special case of that in AutoML, as shown in Figure~\ref{fig:theory}. 
In classical machine learning, the hypothesis space $\mathbf{H}$ is small, because the model usually has little room for adjustment, which makes the approximation error is relatively large. 
For AutoML, we can obtain a larger hypothesis space $\mathbb{H}$ due to the design of the search space through atomization, making approximation error relatively small. 
Moreover, AutoML tries to approach the optimal learning configuration in the search space, which causes the optimization error in AutoML. 
The estimation error in AutoML is similar to that in classical machine learning. 
\end{remark}

From above theoretical analysis, it can be summarized that to improve the performance of AutoML, we can design a larger search space, conduct more iterations of search algorithm or utilize evaluation strategy with more per-iteration cost. 
However, these improvements will all result in more computation cost. 
The trade-off between better AutoML performance and more computation cost will provide guidance for the design of AutoML methods in practice. 

\subsection{Taxonomies of AutoML Approaches}
\label{sec:cate}

\begin{figure}[t]
	\centering
	\includegraphics[width=0.9\textwidth]{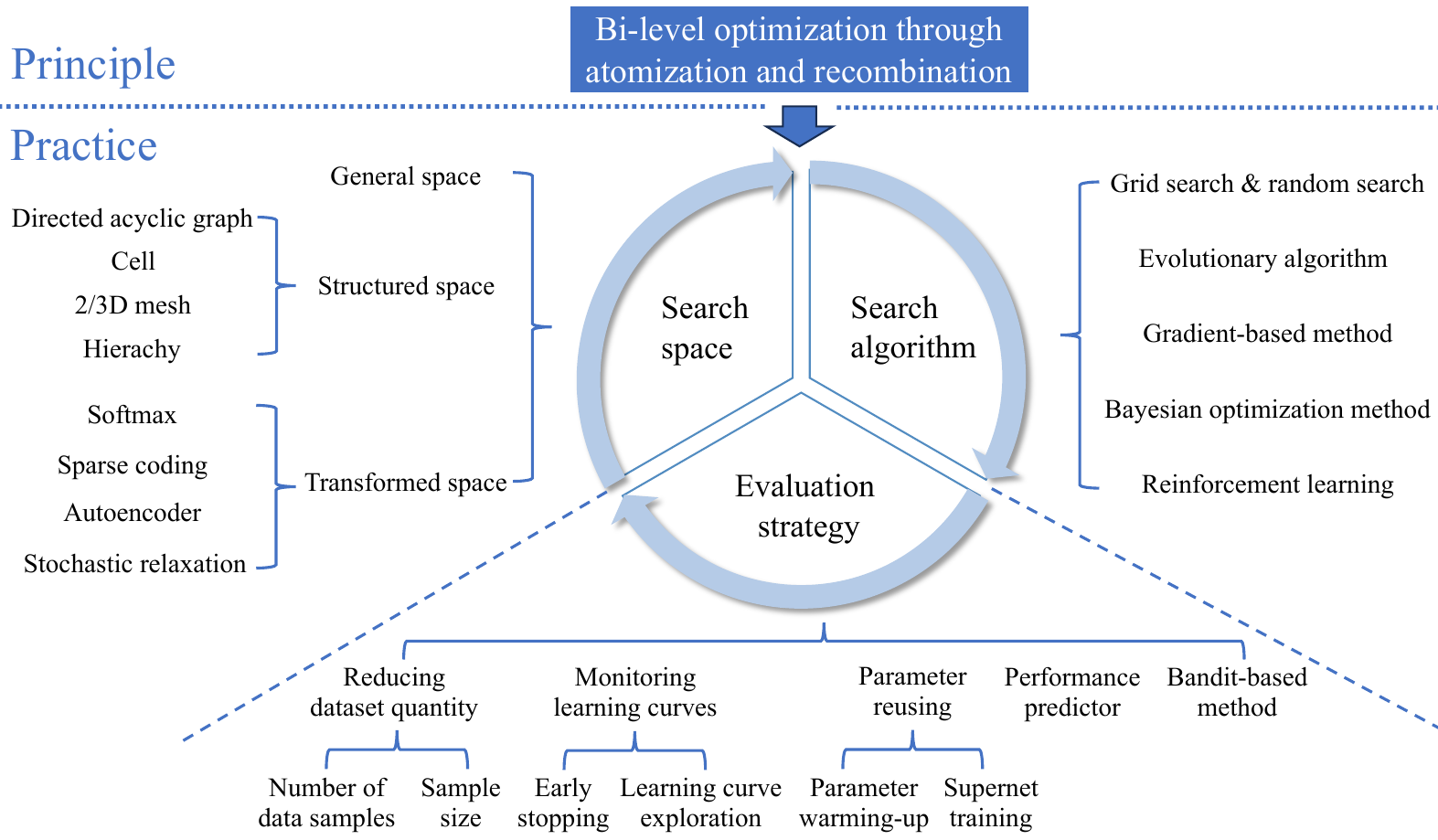}
	\caption{Taxonomies of current AutoML approaches.}
	\label{fig:at_tax}
\end{figure}

In this section, 
we categorize 
existing works in AutoML area by considering the three fundamental factors that researchers have dedicated substantial efforts to address. 
As shown in Figure~\ref{fig:rela_fact}, \textit{search algorithm} 
tries to find good 
learning configurations from the \textit{search space} based on the \textit{evaluation} result of these learning configurations. 
Detailed taxonomies are shown in 
Figure~\ref{fig:at_tax}.

\subsubsection{Search space} 
As shown in Section~\ref{sec:lob_stra}, 
the determination of the search space $\mathcal{A}$ corresponds to the atomization step. 
As in Table~\ref{tab:factor_err}, 
the search space $\mathcal{A}$ should first be large enough to reduce the approximation error $\errapp$. 
Simultaneously, the search space $\mathcal{A}$ should also be compact enough to reduce the computation time, 
lower the optimization error $\erropt$. 
In this survey, we categorize the search space according to its processing stages, 
including general space, 
structured space and transformed space.

\subsubsection{Search algorithm} 

In Figure~\ref{fig:rela_fact}, with search space defined, 
the search algorithm determines how to do recombination 
and finds
a good $\alpha$ from search space $\mathcal{A}$. 
There are two things quantize different search algorithms, i.e., the cost of search process and the quality of search result.
As shown in Table~\ref{tab:factor_err}, 
the search algorithm should conduct 
less iterations,  
but also have precise search result
to reduce optimization error $\erropt$. 
We categorize the optimizer into 
following parts based on the method for candidate generalization: 
grid search and random search, evolutionary algorithm, gradient-based methods, Bayesian optimization method and reinforcement learning.

\subsubsection{Evaluation strategy} 
As shown in Figure~\ref{tab:factor_err}, 
once a specific learning configuration is 
generated from the search space, 
the model parameter for
lower level objective in~\eqref{Eq:bi_automl2} should be 
trained
for configuration evaluation. 
Directly optimizing this lower level objective is usually costly, 
thus various evaluation strategies are proposed to reduce the cost. 
As shown in Table~\ref{tab:factor_err}, the evaluation strategy should be efficient enough to reduce computation time, 
but also provide accurate evaluation for smaller
estimation error $\errest$. 
We categorize the evaluation strategy based on adjustment parts on lower level objective of~\eqref{Eq:bi_automl2}, 
which include reducing data quantity, monitoring learning curves, parameter reusing,  performance predictor and bandit-based methods.

\subsection{Relevant Topics}
\label{sec:rel_topic}

In this section, we discuss several relevant machine learning topics. Their relatedness and difference with respect to AutoML are clarified.

\begin{itemize}[leftmargin=*]
\item \textit{Bi-level optimization}~\cite{dempe2002foundations} is a kind of hierarchical optimization where one problem is embedded within another. 
There are many practical machine learning problems that can be formulated as bi-level optimization problems~\cite{liu2021investigating}, such as meta learning, adversarial learning and deep reinforcement learning. As introduced in Section
~\ref{sec:lob_stra}
, AutoML problems can also be generally regarded as bi-level optimization problems. Therefore, there are many bi-level optimization techniques that are widely applied to handle AutoML problems. 

\item \textit{Meta learning}~\cite{hospedales2021meta}, also viewed as ``learn to learn'', is a technique to improve the performance of machine learning method on new tasks by the provided data set and the meta-knowledge extracted from previous tasks. 
Meta learning has been successfully applied in problems such as few-shot learning~\cite{finn2017model} and reinforcement learning~\cite{flennerhag2021bootstrapped}.
Also, meta learning techniques can be applied to handle real world AutoML problems.
\item \textit{Transfer learning}~\cite{pan2009survey} transfers knowledge from the source domains/tasks, where training data is abundant, to the target domains/tasks, where training data is scarce. Transfer learning can perform well for domain transfer and alleviate the problem of data scarcity in target domain~\cite{ganin2015unsupervised}. 
There are also many examples that transfer learning techniques are applied in AutoML problems. 
\end{itemize}



\section{Search Space}
\label{sec:search_space}

In AutoML, 
the first step is to atomize the classical machine learning solution and define the search space, 
which contains learning
configurations for AutoML solutions of interest. 
Note that to obtain a good search results, the designed search space should be expressive enough to achieve good generalization, while it should also be compact enough to alleviate the burden for searching an optimized configuration from the search space.

There are three search space processing stages: general space, structured space and transformed space.
Firstly, the classical machine learning solution is atomized into a general space $\mathcal{A}$. 
The general space can be structured as a more compacted space $\mathcal{A}'$ to reduce the search space size and alleviate the burden for optimization problem. 
Based on the general space or structured space, we can employ a transformation function $f$ to transform the space into a new space $f(\mathcal{A})$ with better property~(e.g. continuous, differentiable). 
Note that only the atomization step to general space is necessary for AutoML. 
While the processes for structured space and transformed space are optional and they are used to make search space easier to search.


\subsection{General Space}

General space is perhaps the most common search space when using machine learning methods.
The determination of the general search space usually depends on researchers' requirements in specific machine learning problems. 
For example, in the problem of configuring machine learning pipelines, the general search space is usually determined as the choice of important learning configurations~(see search space in Table~\ref{tab:expfactor}).
In the problem of neural architecture search, the general search space is usually the network operation choices and possible connections between operations that researchers focus on. 


With general space determined, a proportion of works in AutoML area~\cite{bergstra2011algorithms, bergstra2012random, li2017hyperband}~(commonly hyperparameter optimization methods) directly set it as the search space. 
For example, the general space may include learning rate ranging from 0.1 to 0.001, batch size ranging from 32 to 256, and optimizer chioces in SGD, Adam and RMSProp.
The general space is usually in the form of Cartesian product and each parameter is set with discrete or continuous choices. 
Other works in AutoML area set the general space as the initial space and then organize or transform it for better optimization, 
which will be introduced in Section~\ref{sec:org} and~\ref{sec:transf}.

\subsection{Structured Space}
\label{sec:org}

Given the general search space $\mathcal{A}$, 
a proportion of works in AutoML area structure the search space into $\mathcal{A}'$ 
to obtain more condense search space. 
Here, 
we mainly categorize the structured search space into four classes: Directed acyclic graph~(DAG), cell, 2/3D mesh and hierarchy. 
Among them, supernet, cell and 2/3D mesh are mainly applied in neural architecture search.  
We summarize the categories with representative works in 
Table~\ref{tab:search_sp_org}. 


\begin{table}[t]
	\footnotesize
	\caption{Representative works for structured space.}
	\vspace{-10px}
	\begin{tabular}{C{35px}|C{150px}|C{170px}}
		\toprule  Category & Method& Description \\ \midrule
		DAG & ENAS~\cite{pham2018efficient}, Few-shot NAS~\cite{zhao2021few}, Primer~\cite{so2021primer} & Regard architecture as different directed acyclic graphs and search for graph structure. \\ \midrule
		Cell &  NAS with RL~\cite{zoph2017neural}, PNAS~\cite{liu2018progressive}, Nas-bench-101~\cite{ying2019bench}, ET~\cite{so2019evolved} & Design cell as basic unit, and search for cell architecture. \\ \midrule
		2/3D mesh & Auto-deeplab~\cite{liu2019auto}, TREFE~\cite{zhang2022searching} & Utilize a 2/3D mesh to search among different learning configuration states. \\ \midrule
		Hierarchy & Mnasnet~\cite{tan2019mnasnet}, GLiT~\cite{chen2021glit}, Auto-WEKA~\cite{thornton2013auto},  AUTO-SKLEARN~\cite{feurer2015efficient} & Design diverse connection among cells and various cell architectures. 
		\\ \bottomrule
	\end{tabular}
	\label{tab:search_sp_org}
\end{table}

\subsubsection{Directed acyclic graph}
\label{sssec:Directed acyclic graph}
\begin{figure}[ht]
	\centering
	\begin{minipage}{0.4\textwidth}
		\centering
		\includegraphics[width=0.95\linewidth]{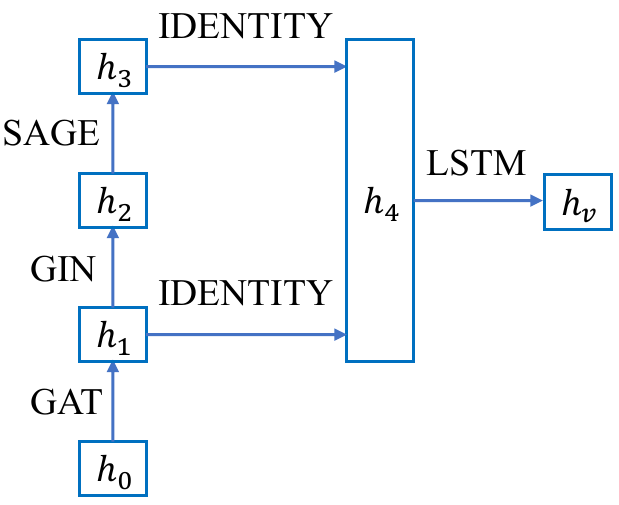}
		\caption{Example for search space structured by directed acyclic graph for GNN.}
		\label{fig:daggnn}
	\end{minipage}
	\quad
	\begin{minipage}{0.5\textwidth}
		\centering
		\includegraphics[width=1.0\linewidth]{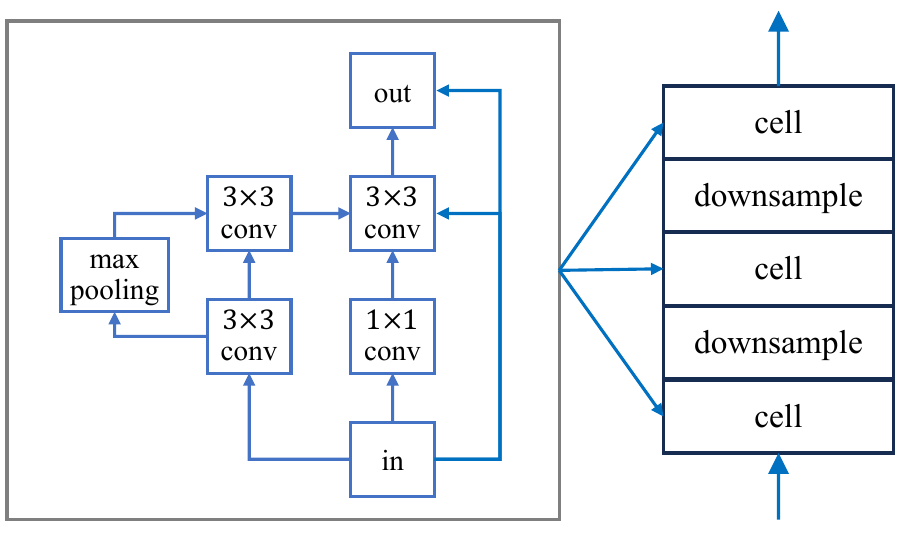}
		\caption{Example for search space structured by cell for CNN.}
		\label{fig:cellcnn}
	\end{minipage}
\end{figure}

Directed acyclic graph~(DAG) is a directed graph with no directed cycles.
An example of search space structured by DAG is shown in 
Figure~\ref{fig:daggnn}. 
In that kind of application, 
we need to search for contents in each node and connection between nodes in the graph. 
When AutoML solution can be generally represented in the form of a flow chart,
the search space structured by DAG can be applied. 
For search space structured by DAG, the most typical application is neural architecture search, 
as the neural architectures can usually be represented as a DAG, 
with input data propagating through the network~(DAG) to output layer. 
The search space structured by DAG is widely used in searching different kinds of neural architectures, including CNN~\cite{pham2018efficient, zhao2021few}
and Transformer~\cite{so2021primer}. 

\subsubsection{Cell} 
Cell is basic structural unit that usually repeatedly appears in a system. 
Figure~\ref{fig:cellcnn} demonstrate the search space structured by cell. 
In that search space, the AutoML solution is designed as connection among cells and the searched cell structure can usually be shared among different cells. 
Search space structured by cell is usually applied when 
there are too much operations to be designed. 
Neural architecture search methods commonly utilize search space structured by cell to compress the search space and better explore prior knowledge on architecture. 
These methods constructed CNN~\cite{zoph2017neural, liu2018progressive, ying2019bench}
, RNN~\cite{klyuchnikov2022bench}, transformers~\cite{so2019evolved} with the repeated cells.

\subsubsection{2/3D mesh}
\begin{figure}[ht]
	\centering
	\begin{minipage}{0.54\textwidth}
		\centering
		\includegraphics[width=0.95\linewidth]{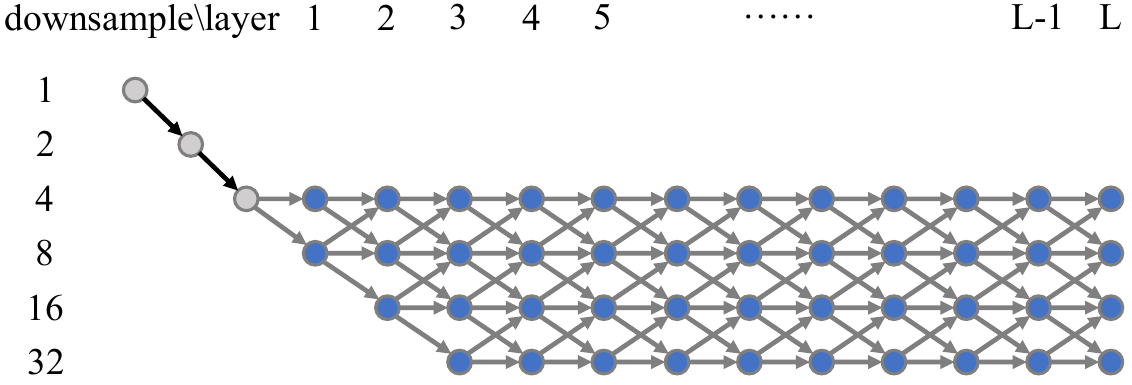}
		\caption{Example for search space structured by 2D mesh for CNN.}
		\label{fig:meshcnn}
	\end{minipage}
	\quad
	\begin{minipage}{0.42\textwidth}
		\centering
		\includegraphics[width=1.0\linewidth]{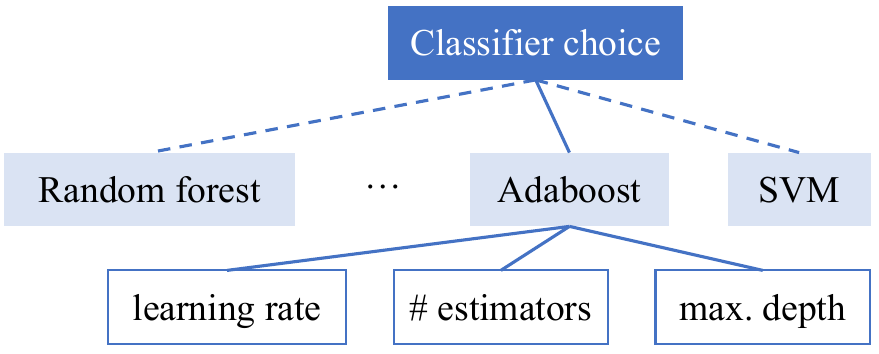}
		\caption{Example for hierarchical search space for machine learning pipeline configuration.}
		\label{fig:hiehpo}
	\end{minipage}
\end{figure}

Mesh is a kind of grid like structure to connect elements. 
Different from other types of search space structures, 
as seen in Figure~\ref{fig:meshcnn}, the search space structured by 2/3D mesh is formed as model for state transitions, where the transition operations we search are on the edges among different states. 
The representative method of this kind of search space is Auto-DeepLab~\cite{liu2019auto}, which searches for the downsampling size~(regarded as state) of each cell and search for the route in the 2D mesh for the best architecture. 

\subsubsection{Hierarchy} 

The hierarchical search space 
is applied when the existence of a part of the search space
is dependent on other operation choices~(see 
Figure~\ref{fig:hiehpo}). 
The usage of hierarchical search space is very extensive. 
In neural architecture search, the cell structure and the connection relation among cells 
can be simultaneously searched~\cite{tan2019mnasnet, chen2021glit}, which form a hierarchical search space. 
In machine learning pipeline configuration, 
a part of hyperparameters exists only when one another specific hyperparameter is selected~\cite{thornton2013auto, feurer2015efficient}
(e.g. the momentum $\beta$ need to be adapted only when optimizer such as RMSprop is selected).

\subsection{Transformed Space}
\label{sec:transf}

In many AutoML problems, the general search space is complex, containing various learning configurations with different value ranges. 
Directly searching in such a complex space will make the optimization problem difficult.
To alleviate this problem,
existing works introduce extra function $f$ to transform the search space $\mathcal{A}$ into a new space $f(\mathcal{A})$ with better property~(e.g. continuous, differentiable, etc.). 
Through transformation, 
the search space will become easier for
search algorithms to operate on.
In this section, the transformation of search space is mainly divided into four classes: softmax, sparse coding, autoencoder and stochastic relaxation. 
Here we summarize the representative works in Table~\ref{tab:search_sp_trans}. 


\begin{table}[t]
	\footnotesize
	\caption{Representative works for transformed search space.}
	\vspace{-10px}
	\begin{tabular}{C{50px}|C{110px}|C{200px}}
		\toprule  Category & Method& Description \\ \midrule
		Softmax & DARTS~\cite{liu2018darts}, PC-DARTS~\cite{xu2019pc}, R-DARTS~\cite{ zela2019understanding} & Use continuous weight parameters to model the importance of configuration choices among candidates. \\ \midrule
		Sparse coding & NASP~\cite{yao2020efficient}, Harmonica~\cite{hazan2018hyperparameter}, ISTA-NAS~\cite{yang2020ista} & Add sparse constraint or set sparse coding objective for the search problem.  \\ \midrule
		Autoencoder & NAO~\cite{luo2018neural}, D-VAE~\cite{zhang2019d}, BRP-NAS~\cite{dudziak2020brp}, GHN~\cite{zhang2018graph} & Map the search space to a latent space with better property for optimization.  \\ \midrule
		Stochastic relaxation & ASNG-NAS~\cite{akimoto2019adaptive}, ENAS~\cite{pham2018efficient}, SNAS~\cite{xie2018snas} & Introduce probability distribution for learning configurations to approximately solve the search problem.  \\ 
		\bottomrule
	\end{tabular}
	\label{tab:search_sp_trans}
\end{table}

\subsubsection{Softmax} 
\label{sssec:transf_softmax}
Softmax is a function commonly used in machine learning to transform a vector of real values into a vector of probabilities.
In AutoML, search space transformed by softmax is to map the weight of discrete operation choices to continuous weight parameters. 
Here we take search space transformation of DARTS~\cite{liu2018darts} as an example. 
Denote $\mathcal{O}$ as a set of candidate operations for a certain module and each operation represents function $o(\cdot)$ to be applied to input $x$. 
Then the output is calculated as:
\begin{align}
\begin{split}
\label{Eq:trans_darts}
\bar{o}(x) = \sum \limits_{o\in \mathcal{O}} \frac{exp(\alpha_o)}{\sum\nolimits_{o' \in \mathcal{O}} exp(\alpha_{o'}) } o(x),
\end{split}
\end{align}
where $\alpha_o$ is the weight parameters corresponding to candidate operations. 
Through transformation by softmax, parameters of all operations in $\mathcal{O}$ and weight parameters $\alpha_o$ can be trained together, which enhance the training efficiency. 
Similar to DARTS, several works in AutoML area~\cite{xu2019pc, zela2019understanding} also conduct search space transformed by softmax to achieve supernet training. 

\subsubsection{Sparse coding} 
Sparse coding~\cite{donoho2006compressed}
is a representation learning method that aims at finding a sparse representation of the input data in the form of a linear combination of basic elements. 
In AutoML, search space transformed by sparse coding is to extract key components from the search space~(usually denote as a dictionary $B = \{\beta_1, ..., \beta_n \}$) and then transform learning configurations $\alpha$ in the search space into sparse representations by components in $B$~(as shown in~\eqref{Eq:sp_coding}). 
\begin{align}
\label{Eq:sp_coding}
c^* = \argmax\nolimits_{c} ||\alpha - \sum\nolimits_{i=1}^{n} c_i \beta_i||,  
\quad\text{s.t.}\quad 
\sum\nolimits_{i=1}^{n} c_i = 1, c_i \geq 0, \forall i \in \{1, ..., n\}, 
\end{align}
Generally, transformation by sparse coding largely relies on the prior knowledge on the search space, 
through which critical parts in the search space can be identified and the search space can be compressed. 
For example, NASP~\cite{yao2020efficient} remove redundant connections in the neural network and add sparsity constraints to the weight parameters of the network based on the prior knowledge that only one operation in each module is selected in optimized model. 
Related AutoML works~\cite{hazan2018hyperparameter, yang2020ista} have been proposed to utilize sparse coding to transform the search space.

\subsubsection{Autoencoder} 

\begin{figure}[htbp]
\centering
\includegraphics[width=0.95\textwidth]{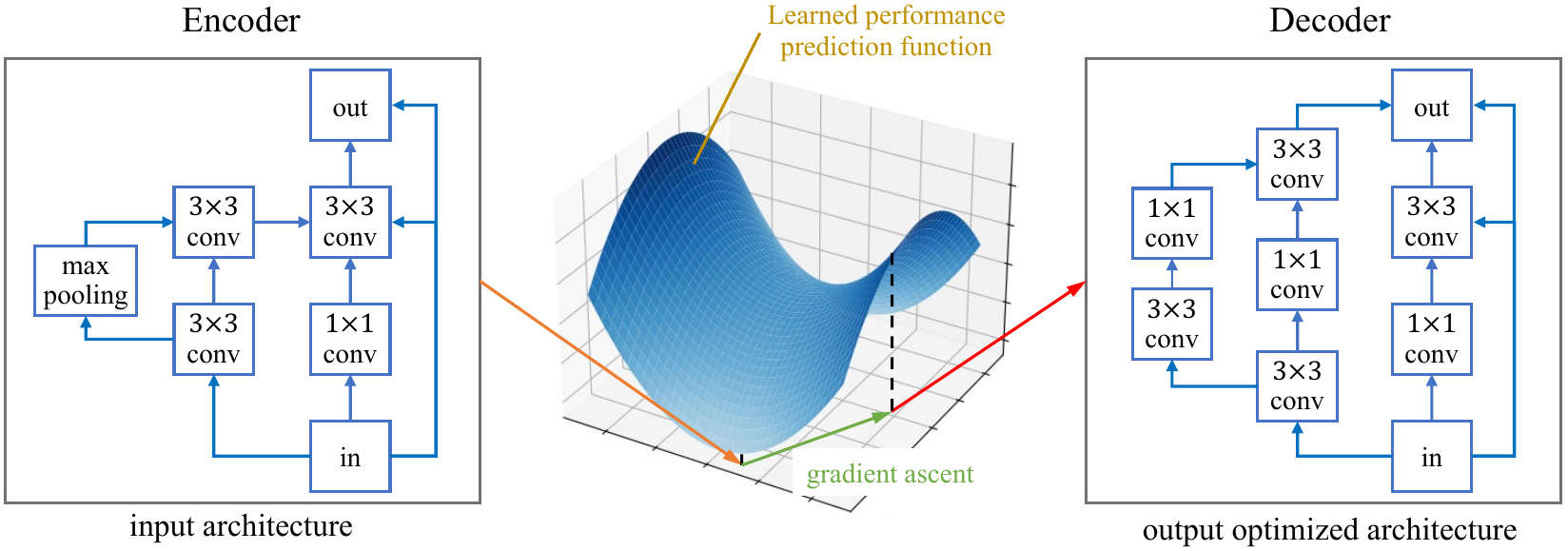}
\caption{The framework of transformed space by autoencoder~\cite{luo2018neural}. 
}
\label{fig:nao_sp}
\end{figure}

Autoencoder 
is a kind of artificial neural network used to learn efficient coding and the encoding space should possess useful properties. 
It contains an encoder to transform data into the encoding space and a decoder to reconstruct the data from the encoding space. 
In AutoML, search space transformed by autoencoder is able to transform complex search space into a continuous space to make the optimization problem easier. 
Its framework is shown in Figure~\ref{fig:nao_sp}.
For each set of learning configurations~(architecture), an autoencoder is first trained to map it to a continuous space. 
Then optimization method can be utilized to find a good point in that space. 
Finally, the best set of learning configurations~(architecture) can be obtained by a decoderfrom that point in the continuous space. 
There exists search space encoding methods~\cite{zhang2019d, dudziak2020brp, zhang2018graph} that utilize various models as both encoders and decoders, 
such as Multilayer Perceptron~(MLP) and Graph Neural Networks~(GNNs).

\subsubsection{Stochastic relaxation}

Stochastic relaxation 
is a technique that allows optimizing non-differentiable functions by transforming them into differentiable alternatives.
In AutoML, the complex optimization problem refers to searching for a good learning configuration in the 
search space~(see~\eqref{Eq:bi_automl1}~\eqref{Eq:bi_automl2}) and search space transformed by stochastic relaxation maps each search space configuration 
$\alpha$ into a probability distribution $p_\theta(\alpha)$ with distribution parameter $\theta$. 
With that distribution, we can obtain the learning objective $P$ as follows:
\begin{align}
\begin{split}
P(\mathbf{x},\theta) = \int_{\alpha} f(\mathbf{x}, \alpha) p_\theta(\alpha) d\theta = \mathbb{E}_{p_\theta}[f(\mathbf{x}, \alpha)],
\end{split}
\label{eq:sto_relax}
\end{align}
where $f$ is the objective function. 
In~\eqref{eq:sto_relax}, the non-differentiable configuration $\alpha$ no longer need to be considered, 
and its optimization can be replaced by the optimization of differentiable distribution parameter $\theta$. 
There exist works in AutoML area~\cite{pham2018efficient, akimoto2019adaptive, xie2018snas} that make use of the idea of stochastic relaxation for search space transformation to handle the search problem. 

\section{Search Algorithm}
\label{sec:search_algorithm}

In AutoML, the main function of search algorithm is to find 
a learning configuration in the search space that has good performance~(i.e. the upper level objective in~\eqref{Eq:bi_automl1}). 
To design a good AutoML search algorithm, the following trade-off should be taken into account:
efficient searching methods usually cannot guarantee the quality of the search result, while searching method with good convergence property is usually time-consuming. 
Based on the choice of candidate generation function, we categorize the search algorithm into grid search and random search, evolutionary algorithm, gradient-based methods, Bayesian method and reinforcement learning.
We summarize the typical types of search algorithms, their representative works and characteristics in 
Table~\ref{tab:search_alg}.

\begin{table}[t]
	\footnotesize
	\caption{Representative search algorithms in AutoML.}
	\vspace{-10px}
	\begin{tabular}{ C{51px}|C{85px}|C{110px}|C{100px}}
		\toprule
		Category & Method & Candidate generation function & Usage requirement \\ \midrule
		Grid search and random search & Random Search~\cite{bergstra2012random}, GSHPT~\cite{shekar2019grid} & Select in a fixed order / select randomly & Model-free, no special requirements \\ \midrule
		Evolutionary algorithm &  TPOT~\cite{olson2016evaluation}, EVOLUTION~\cite{real2017large}, ET~\cite{so2019evolved}, MENNDL~\cite{young2015optimizing} & Mutation and crossover from candidate learning configuration set & 
		Model-free, no special requirements \\ \midrule
		Gradient-based method &  DARTS~\cite{liu2018darts}, ASNG-NAS\cite{akimoto2019adaptive}, S2E~\cite{yao2020searching}, ENAS~\cite{pham2018efficient} & Select in the direction of gradient & 
		Model-free, evaluation metric differentiable \\ \midrule
		Bayesian optimization method & Arc GP~\cite{swersky2014raiders}, NASBOT~\cite{kandasamy2018neural}, TPE~\cite{bergstra2011algorithms}, BOHB~\cite{falkner2018bohb} & Learning configuration that prone to have good performance according to the surrogate model & Model-based, evaluation metric can be fit by probabilistic model \\ \midrule
		Reinforcement learning & NAS with RL~\cite{zoph2017neural}, MetaQNN~\cite{baker2017designing}, NASNet~\cite{zoph2017learning} & Select based on a designed controller & 
		Model-based, evaluation metric can provide effective information for the controller \\
		\bottomrule
	\end{tabular}
	\label{tab:search_alg}
\end{table}

\subsection{Grid search and random search}
\begin{figure}[t]
	\centering
	\includegraphics[width=0.6\textwidth]{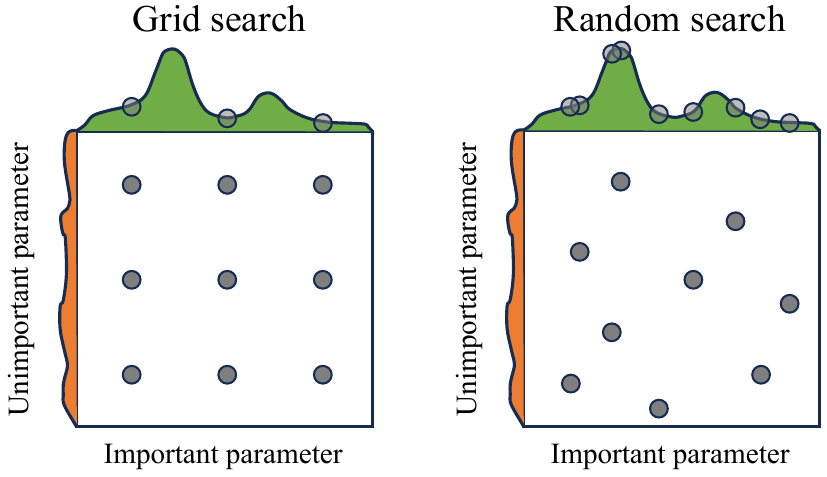}
	\caption{Illustration of grid search and random search in a simple search problem~\cite{bergstra2012random}. 
		}
	\label{fig:random_search}
\end{figure}

Grid search is 
the simplest hyperparameter optimization method, 
which exhaustively searches for all possible 
learning configurations following a grid-liked pattern in the search space. 
Random search~\cite{bergstra2012random} randomly sample learning configurations in the search space and evaluate their performance to find the best configuration. 
It has been empirically demonstrated that random search performs better than grid search, because random search usually has a better exploration on important learning configurations~(see Figure~\ref{fig:random_search}). 
Grid search and random search are usually used as baselines of other newly proposed AutoML search algorithms~\cite{thornton2013auto, liu2018darts}. 
These methods requires no prior knowledge while ignore the previous evaluations, which largely limit their search efficiency. 

\subsection{Evolutionary algorithm}
\label{sec:evl_alg}

\begin{figure}[htbp]
	\centering
	\includegraphics[width=0.9\textwidth]{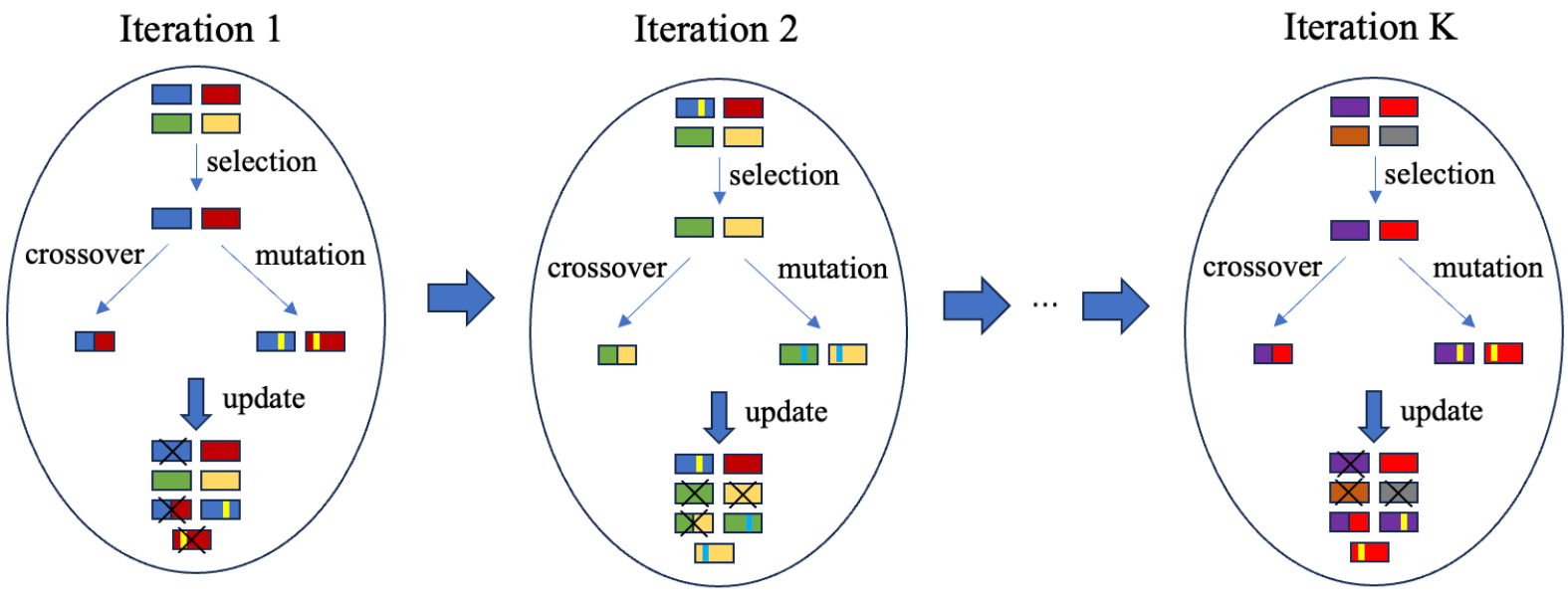}
	\caption{Illustration of evolutionary algorithm procedure. }
	\label{fig:evo}
\end{figure}

Evolutionary algorithm 
is a kind of population based search algorithm that draws inspiration from the process of natural evolution. 
For evolutionary algorithms in AutoML, a set of candidate learning configurations are usually maintained. 
In each step, several kinds of operations can be conducted to update the candidate learning configuration set~(Figure~\ref{fig:evo}): selection, which select parent learning configurations 
for later evolution steps;
crossover, 
exchange the parts of two parent learning configurations; mutation, which randomly change a configuration to generate a new one; mutation, which randomly change a configuration slightly; 
update, which add newly generated learning configurations and remove the learning configurations with unsatisfactory performance from the set of candidate learning configurations. 
Running the above operations iteratively, the candidate configuration set will be 
updated for better performance. 
The evolutionary algorithm is widely applied to handle complex AutoML problems, such as automated feature engineering~\cite{vafaie1992genetic, smith2005genetic, olson2016evaluation}, neural architecture search~\cite{real2017large, real2019regularized, so2019evolved} and hyper-parameter optimization~\cite{aszemi2019hyperparameter, young2015optimizing}. 

\subsection{Gradient-based method}
\label{sec: grad method}
Gradient-based method,
such as gradient descent,
is commonly used to solve optimization problems. 
In AutoML, if the gradient of validation evaluation metric w.r.t. learning configuration can be obtained, 
candidate learning configurations can be generated by
following the direction of gradient. 
Theoretically, the gradient of the performance w.r.t. learning configurations can be calculated as follows~\cite{bengio2000gradient}: 
\begin{align}
	\begin{split}
	\label{Eq:gradient}
	\frac{P(C_{\alpha}(w^*), \mathcal{D}_{\text{valid}})}{\partial \alpha}  = \frac{\partial P(C_{\alpha}(w^*), \mathcal{D}_{\text{valid}})}{\partial w^*} \frac{\partial w^*}{\partial \alpha} + \frac{\partial P(C_{\alpha}(w^*), \mathcal{D}_{\text{valid}})}{\partial \alpha},\\
	\text{where} \quad \frac{\partial w^*}{\partial \alpha} = - (\frac{\partial^2 L(C_{\alpha}(w^*), \mathcal{D}_{\text{train}})}{\partial {w^*}^2})^{-1} \frac{\partial^2 L(C_{\alpha}(w^*), \mathcal{D}_{\text{train}})}{\partial w^* \partial \alpha}.
	\end{split}
\end{align}
However, in real-world AutoML problems, it is usually difficult to obtain the second-order derivative of the optimized model parameter $w^*$ w.r.t. learning configuration parameter. 
To estimate the gradient, common strategies can be mainly divided into two classes: 
iterative differentiation and approximate implicit differentiation. 

Iterative differentiation is a common strategy to calculate gradient in AutoML problems. 
When calculating the gradient of validation performance w.r.t. learning configurations,
the iterative differentiation methods first iteratively update the model parameter for n steps to obtain $w_n(\alpha)$ for the optimization of lower level objective, and then uses the updated result $w_n(\alpha)$ to approximate $w^*(\alpha)$ to obtain the approximated gradient.
As the model training for the lower level objective is usually considered as a dynamic system, 
the gradient of validation evaluation metric $P$ w.r.t. learning configuration $\alpha$ can be obtained through this process. 
For example, many gradient based methods for neural architecture search methods~\cite{liu2018darts, chen2019progressive, chen2020stabilizing} utilize iterative differentiation to obtain the gradient of architecture weight parameters. 
Based on iterative differentiation, different kinds of gradient-related techniques are also utilized.
Proximal gradient~\cite{yao2020efficient, wu2021neural} is employed to hasten the convergence rate. 
Additionally, natural gradient~\cite{akimoto2019adaptive} is utilized to enhance both the speed and quality of optimization convergence. 

Different from iterative differentiation, the approximate implicit differentiation tries to directly utilize results in~\eqref{Eq:gradient} and approximate the second-order derivative. 
In several hyper-parameter optimization methods, the idea of approximate implicit differentiation is utilized to obtain the approximate gradient of hyper-parameters~\cite{pedregosa2016hyper, lorraine2020optimizing}. 
However, in real-world AutoML problems, it is usually hard to obtain $\partial w^* / \partial \alpha$ through the explicit expression of the gradient, which limits the application of approximate implicit differentiation.


\subsection{Bayesian Optimization method}
\label{sec:alg_bay}

\begin{figure}[ht]
	\centering
	\begin{minipage}{0.44\textwidth}
		\centering
		\includegraphics[width=0.95\linewidth]{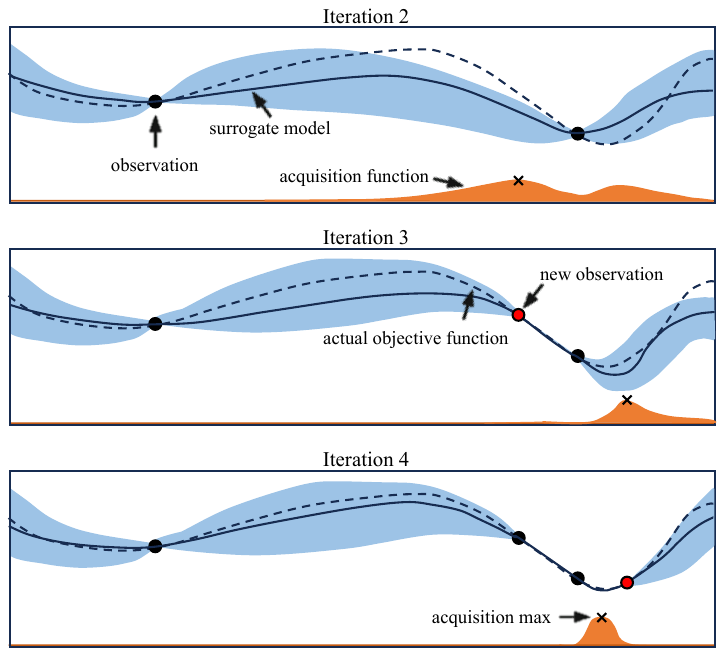}
		\caption{Bayesian optimization method~\cite{vincent2023improved}. }
		\label{fig:bay}
	\end{minipage}
	\quad
	\begin{minipage}{0.52\textwidth}
		\centering
		\includegraphics[width=1.0\linewidth]{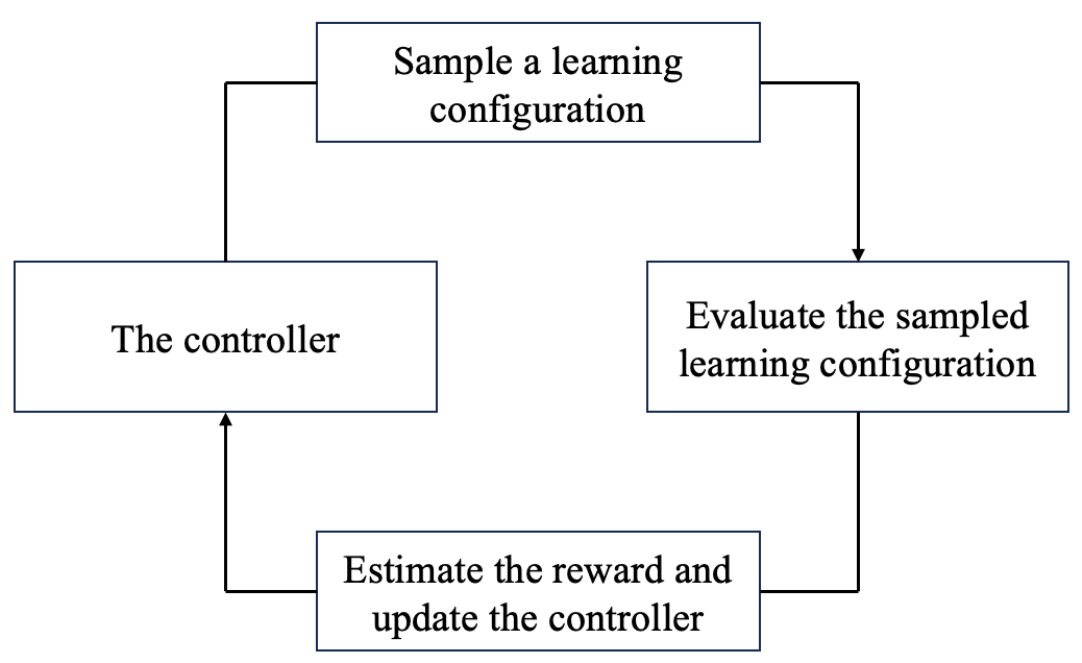}
		\caption{The algorithm for reinforcement learning. }
		\label{fig:alg_rein}
	\end{minipage}
\end{figure}

Bayesian optimization 
is a sequential design strategy for global optimization of black-box functions that does not assume any functional forms. 
In AutoML, Bayesian optimization method mainly involves two parts of design: 
a probabilistic surrogate to model the relation between learning configuration and its corresponding performance; 
and an acquisition function to decide the next configuration to be evaluated depend on the surrogate model. 
When designing these two parts, the surrogate model should be flexible to fit the observations, and the acquisition function should be able to balance the exploration and exploitation of searching.
When conducting search process, 
Bayesian optimization method adapts the surrogate model and acquisition function with new observations of evaluation results~(Figure~\ref{fig:bay}).
As Bayesian optimization method do not add much assumption on the search space, it can be applied to various kinds of AutoML problems, such as neural architecture search~\cite{swersky2014raiders, kandasamy2018neural}, hyper-parameter optimization~\cite{bergstra2011algorithms, falkner2018bohb}. 

\subsection{Reinforcement learning}
Reinforcement learning 
is a subset of machine learning that allows an agent to learn through trial and error using feedback from its actions. 
The AutoML problem can be formed as a reinforcement learning problem, 
because the search process in AutoML is actually a trial and error process based on the feedback of evaluation results for different learning configurations. 
The framework of reinforcement learning used in AutoML is shown in Figure~\ref{fig:alg_rein}. 
The controller is usually used to sample candidate learning configurations 
based on the evaluation performance of the configurations sampled before. 
NASNet~\cite{zoph2017neural} is the first AutoML method that utilizes reinforcement learning to handle neural architecture search. 
Following similar strategy, ENAS~\cite{pham2018efficient} largely improve the efficiency using weight sharing. 
Zoph et. al. \cite{zoph2017learning} designs a new search space to enable better transfer ability among different datasets.

\section{Evaluation Strategy}
\label{sec:evaluation}
In the problem formulation of AutoML at Section~\ref{sec:lob_stra}, 
the function of evaluation is to obtain the performance of a specific learning configuration, 
which is usually served as a feedback for the search algorithm to find better candidate learning configurations. 
The most accurate evaluation is to train the model from scratch to convergence, but it is too time-consuming.  
In contrast, fast evaluation methods are able to return the performance of a learning configuration in a short time, 
but the evaluation result is usually inaccurate.
Therefore, when designing evaluation methods, 
the trade-off between evaluation speed and accuracy should be considered. 
According to the adjustment on lower level objective in~\eqref{Eq:bi_automl2}, the evaluation acceleration can be divided into the following categories:
reducing dataset quantity of $\mathcal{D}_{\text{train}}$, monitoring learning curves when conducting $\argmax$, reusing parameters $w$ among different learning configurations and designing a surrogate model for instead of the initial performance measurement $P$.
Here, we also summarize the typical evaluation methods with representative works in Table~\ref{tab:search_eva}.

\begin{table}[t]
	\footnotesize
	\caption{Representative works for AutoML on evaluation.}
	\vspace{-10px}
	\begin{tabular}{ C{35px}|C{75px}|C{100px}|C{120px}}
		\toprule  \multicolumn{2}{c|}{Category} & Method & Adjustment on lower level objective of~\eqref{def:automl} \\ \midrule
		\multirow{3}{*}{\makecell[c]{Reducing \\ dataset \\ quantity}} &  Reducing the number of
		data samples & TSE~\cite{hu2019multi}, GTN-NAS~\cite{such2020generative}  & \multirow{3}{*}{\makecell[c]{Reduce the quantity of dataset $\mathcal{D}_{\text{train}}$}} \\ \cmidrule{2-3}
		&Reducing sample size & Downsampling~\cite{chrabaszcz2017downsampled}, Petridish~\cite{hu2019efficient}  &  \\\midrule
		\multirow{3}{*}{\makecell[c]{Monitoring \\ learning \\ curves }} & Early stopping & Hyperband~\cite{li2017hyperband}, BOHB~\cite{falkner2018bohb}, DARTS+~\cite{liang2019darts+} & \multirow{3}{*}{\makecell[c]{Monitor and accelerate the \\ process of $\argmax$}} \\ \cmidrule{2-3}
		&Learning curve exploration & \cite{domhan2015speeding}, LC~\cite{klein2016learning}, \cite{baker2017accelerating} & \\ \midrule
		\multirow{2}{*}{\makecell[c]{Parameter \\ reusing} } & Parameter warming-up &  EXPSRACOS~\cite{hu2018experienced}, model-reuse~\cite{bajcsy2022characterization} & \multirow{3}{*}{\makecell[c]{Give the model parameter $w$ \\ a better initialization}}  \\ \cmidrule{2-3}
		& Supernet training & ENAS~\cite{pham2018efficient}, DARTS~\cite{liu2018darts} & \\ \midrule
		\multicolumn{2}{c|}{Performance predictor} & Surrogate benchmarks~\cite{eggensperger2015efficient}, PNAS~\cite{liu2018progressive}, zero-cost NAS~\cite{abdelfattah2020zero} & Use a performance predictor as a substitute of $P$ \\ \midrule
		\multicolumn{2}{c|}{Bandit-based method} & SuccessiveHalving~\cite{jamieson2016non}, Hyperband~\cite{li2017hyperband}, BOHB~\cite{falkner2018bohb} & Reduce the quantity of dataset $\mathcal{D}_{\text{train}}$, or accelerate the process of $\argmax$ \\ 
		\bottomrule
	\end{tabular}
	\label{tab:search_eva}
\end{table}

\subsection{Reducing dataset quantity}
\label{sec:reduce_qd}

\begin{figure}[t]
	\centering
	\includegraphics[width=0.8\textwidth]{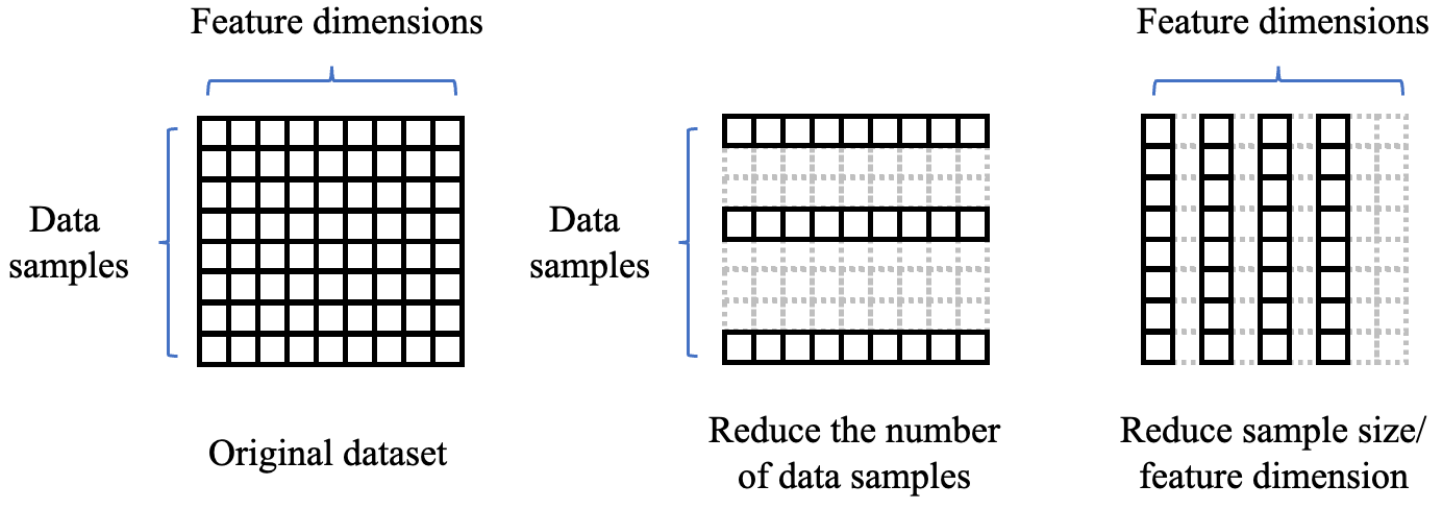}
	\caption{Illustration of two ways to reduce data quantity. }
	\label{fig:reduce_data}
\end{figure}

As shown in Section~\ref{sec:theory}, 
the training objective in machine learning is usually expressed as
a summation of loss over a large set samples. 
Therefore, an intuitive perspective to reduce evaluation cost is to reduce dataset quantity, 
so that the training process can be accelerated.
Dataset quantity can be divided into two categories: 
the number of data samples and 
the sample size or feature dimension~(Figure~\ref{fig:reduce_data}).
For reducing the number of data samples, 
several works consider to sample a subset of the dataset~\cite{hu2019multi, na2021accelerating, killamsetty2022automata, prasad2022speeding}. 
It can be also achieved through the generation of synthetic data~\cite{such2020generative} to represent the original dataset. 
Reducing data sample size or feature dimension is also a common way to reduce dataset quantity. 
It can be accomplished by downsampling~\cite{chrabaszcz2017downsampled, zogaj2021doing}~(mainly for images) 
or feature selection~\cite{chandrashekar2014survey, hu2019efficient}. 
Generally, the less data is used, the faster the evaluation will be, but the expectation of evaluation error will be higher. 

\subsection{Monitoring learning curves}
\label{sec:eva_lc}

\begin{figure}[htbp]
	\centering
	\includegraphics[width=0.95\textwidth]{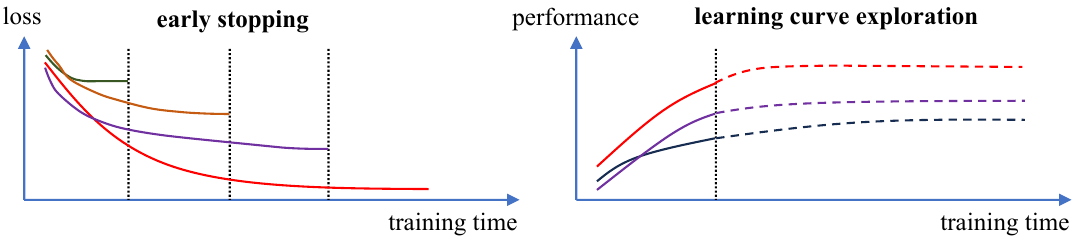}
	\caption{Illustration of early stopping and learning curve exploration in monitoring learning curves. }
	\label{fig:monitor_curve}
\end{figure}

Monitoring learning curves mainly consider to monitor the training 
step~(i.e. the $\argmax$ step in the lower level objective of~\eqref{Eq:bi_automl2}). 
It can be mainly divided into two categories: early stopping and learning curve exploration~(Figure~\ref{fig:monitor_curve}).
In classical machine learning, early stopping is mainly used to avoid overfitting of the model~\cite{goodfellow2016deep}. 
In AutoML, apart from avoiding overfitting, 
early stopping is used to reduce the evaluation cost, 
which terminate the training process of a proportion of learning configurations in advance when their performance in early training stage is poor or the performance does not get improved
for a certain period~\cite{li2017hyperband, falkner2018bohb, liang2019darts+}. 
Learning curve exploration tries to predict the performance based on partial learning curves. 
The prediction can be achieved by traditional curve fitting methods~\cite{domhan2015speeding} or training a surrogate model for prediction~\cite{klein2016learning, baker2017accelerating}.

\subsection{Parameter reusing}
\label{sec:para_re}
\begin{figure}[htbp]
\centering
\includegraphics[width=0.95\textwidth]{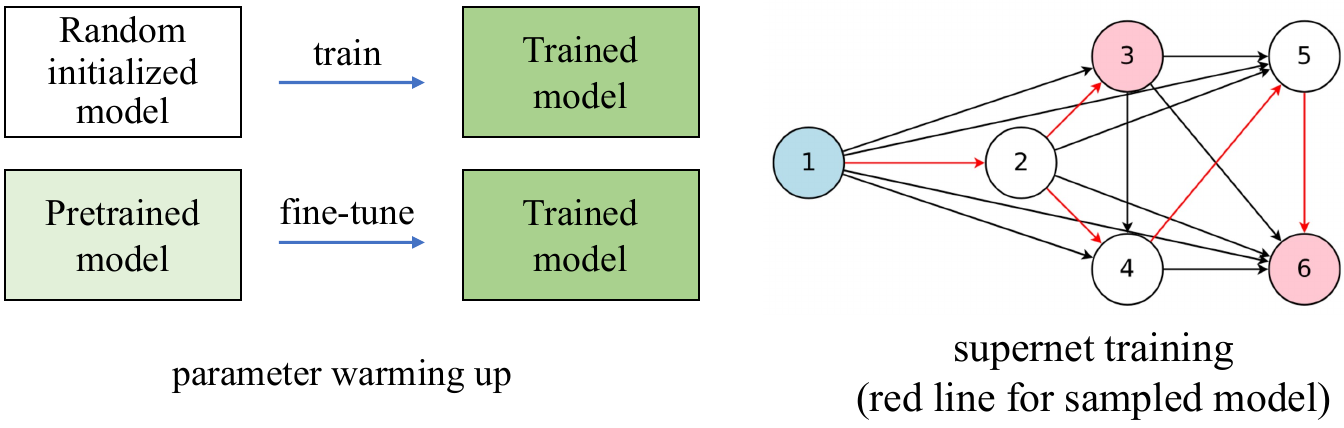}
\caption{Illustration of parameter warming-up and supernet training~\cite{pham2018efficient} in parameter reusing. }
\label{fig:para_reuse}
\end{figure}

Parameter reusing mainly consider to give the model parameter $w$ in~\eqref{def:automl} a better initialization, 
so that the evaluation process can be accelerated. 
It can be mainly divided into parameter warming-up and supernet training~(Figure~\ref{fig:para_reuse}). 
Parameter warming-up first collects the model trained in other datasets or trained under other configurations as the initialization and then
fine-tune the model in the current settings~\cite{hu2018experienced, bajcsy2022characterization}, which can accelerate the convergence speed of the model. 
Generally, the pre-training setting of the initialzed model should be similar to the current setting. 

Supernet training is another parameter reusing technique that is usually applied in neural architecture search. 
For neural architecture search problems, each learning configuration usually corresponds to a neural network, and different neural networks share a proportion of components. 
Based on this, a supernet can be constructed, which contains all the possible neural networks in the search space. 
Then the supernet can be trained once and the parameters of the supernet can be reused to evaluate the performance of different learning configurations~\cite{pham2018efficient,liu2018darts}~(e.g. using parameter weights according to index in the figure above), which can largely reduce the evaluation cost.
Nevertheless, the supernet may have the problems when the performance of shared model parameters fails to accurately reflect the actual performance of the corresponding learning configuration~\cite{li2020geometry}. 
This discrepancy can lead to significant evaluation errors. 
Also, the supernet is usually much larger than a normal neural network, which means supernet training has high memory cost.

\subsection{Performance predictor}
\label{sec:perf_pd}

\begin{figure}[htbp]
	\centering
	\includegraphics[width=0.8\textwidth]{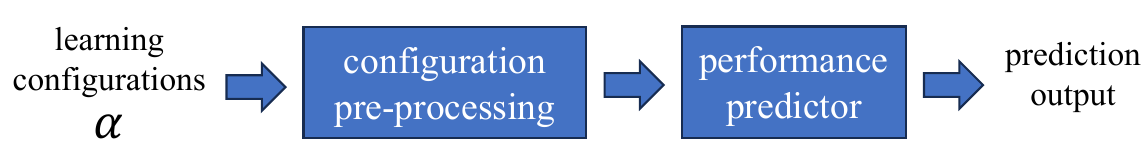}
	\caption{Illustration of performance predictor. }
	\label{fig:eva_sur}
\end{figure}

Performance predictor uses a surrogate model to predict the performance of learning configurations, 
which is a substitute of the performance measurement $P$ in Eq.~\eqref{def:automl}. 
As shown in Figure~\ref{fig:eva_sur}, the performance predictor can be mainly divided into two steps: learning configuration pre-processing and performance prediction. 
The main function of configuration pre-processing is to collect the feature of the learning configurations without conducting the full training procedure. 
The features utilized by different methods can be various, which include the property of the constructed model~\cite{liu2018progressive, lin2021zen, abdelfattah2020zero} and the output of the untrained model with different data input~\cite{mellor2021neural, ozturk2022zero}. 
With these features as input, the designs of performance predictors are also diverse, which include hand-designed prediction function~\cite{mellor2021neural, lin2021zen}, CNN~\cite{ozturk2022zero}, RNN~\cite{liu2018progressive} and so on. 
The surrogate model of the performance predictor is usually much smaller than the model to be evaluated, so the evaluation process can be faster than training the original model from scratch.
However, the use of performance predictor has several prerequisites. 
The effectiveness of the performance predictor lies in its ability to learn the correlation between learning configurations and their performance, enabling it to provide rough estimates of performance for learning configurations.
Also, the correlation between configurations and performances can be complex, such that simple predictors can not capture the correlation, while training a complex predictor requires much supervised data samples. 
Note 
that the learning curve predictor~\cite{domhan2015speeding, klein2016learning, baker2017accelerating}~(in Section~\ref{sec:eva_lc}) and
the surrogate model in Bayesian optimization~(in Section~\ref{sec:alg_bay}) are  performance predictors as well. 

\subsection{Bandit-based method}
\label{sec:bandit}
In classical machine learning, the bandit problem involves optimally allocating a fixed set of resources among competing choices with partially known properties, aiming to maximize expected gain. 
In AutoML, the evaluation acceleration problem can be regarded as a bandit problem, 
where the limited resources are given as the evaluation budget and competing choices correspond to different learning configurations in the search space. 
In that problem, 
bandit-based methods are able to evaluate more learning configurations by allocating different kinds of resources, 
e.g., allocating time by early stopping or cutting cost by reducing data quantity~\cite{li2017hyperband}. 
SuccessiveHalving~\cite{jamieson2016non} 
first uniformly allocates training resources on a large number of learning configurations, then iteratively eliminates 
half of the less promising training processes and uniformly allocate training resources on remaining parts.
For example, hyperband first uniformly allocates the training resources on 32 learning configurations, then kills off 16 configurations that are not promising 
and uniformly allocates the training resources on remaining 16 learning configurations, and so on until only 1 configuration remains.
Hyperband~\cite{li2017hyperband} makes improvement by repeatedly calling SuccessiveHalving and control the speed of training termination, which better balances the exploration and exploitation of learning configuration evaluation. 
Bayesian optimization Hyperband~(BOHB)~\cite{falkner2018bohb} further trains a probabilistic model to choose the learning configurations in the termination step of Hyperband algorithm. 


\section{Exemplary Applications}
\label{sec:application}

\subsection{Automated Configuring Machine Learning~(ML) Pipeline}

\begin{figure}[t]
\centering
\includegraphics[width=0.85\textwidth]{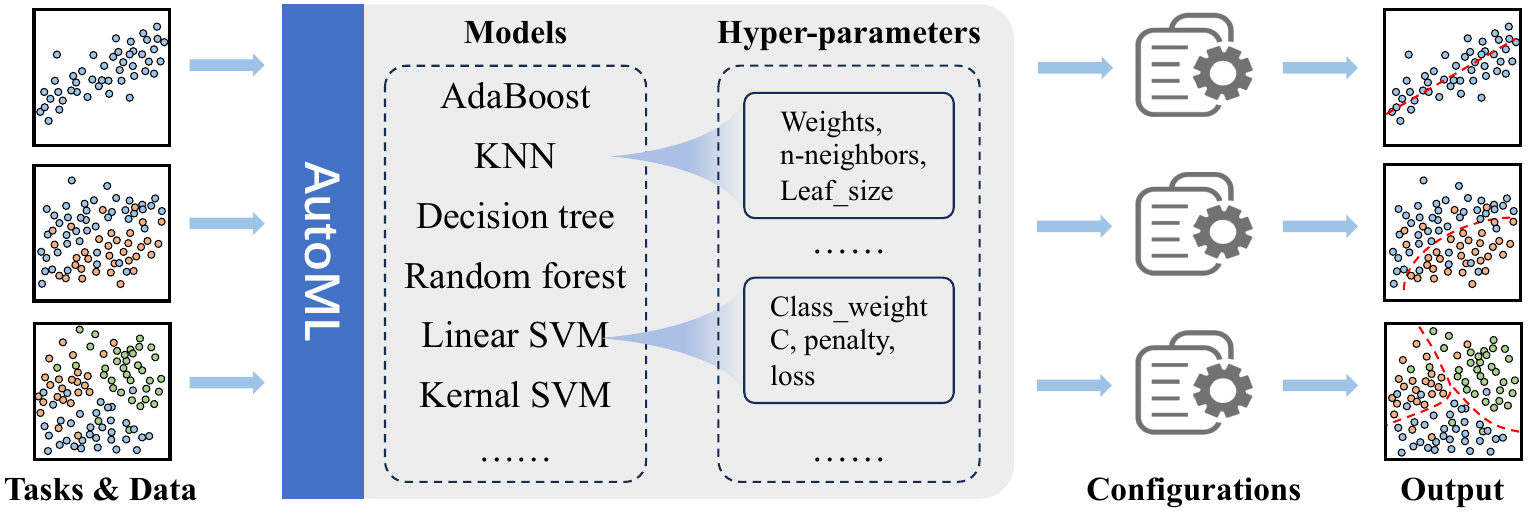}
\caption{AutoML can get involved in the ML pipeline and minimize participation of humans.}
\label{fig:ML_config}
\end{figure}

The set of ML pipeline configurations is a structured framework outlining the complete workflow in an ML project. 
This workflow includes stages like
feature engineering, model and algorithm selection, hyper-parameter optimization and evaluation. 
Since each task has its own preference over learning tools~\cite{tom1997machine}, 
suitable machine learning pipeline configurations should be chosen specifically. 
AutoML can significantly reduce the need for manual intervention in configuring the pipeline. 
This automation is particularly beneficial for non-expert users, 
enabling them to efficiently leverage ML solutions for their practical challenges without requiring deep technical knowledge. 
Here, 
we show how AutoML can automatically find the satisfactory set of ML pipeline configurations   (as shown in Figure \ref{fig:ML_config}).

The issue of finding the optimal set of ML pipeline configurations has been formulated as a Combined Algorithm Selection and Hyper-parameter (CASH) problem (Example \ref{def:cashpro}) by previous AutoML techniques and frameworks~\cite{thornton2013auto, feurer2015efficient, feurer2022auto},
which aims to minimize the validation loss with respect to the model as well as its hyper-parameters and parameters.


\vspace{5px}
\begin{example}[CASH Problem~\cite{thornton2013auto,feurer2015efficient,hutter2019automated}]
\label{def:cashpro}
Let $\mathcal{F} = \{ F_1, \cdots, F_R \}$ be a set of learning models,
and each model has hyper-parameter $\theta_j$ with domain $\Lambda_j$, 
$\mathcal{D}_{\text{train}} = \{ \left( \vect{x}_1, y_1 \right), \cdots, \left( \vect{x}_n, y_n \right) \}$
be a training set which is split into $K$ cross-validation folds
$\{ \mathcal{D}_{\text{train}}^{1}, \cdots, \mathcal{D}_{\text{train}}^{K} \}$
and
$\{ \mathcal{D}_{\text{valid}}^{1}, \cdots, \mathcal{D}_{\text{valid}}^{K} \}$
with $\mathcal{D}_{\text{train}}^{i} \cup \mathcal{D}_{\text{valid}}^{i} = \mathcal{D}_{\text{train}}$
for $i = 1, \dots, K$.
Then,
the Combined Algorithm Selection and Hyper-parameter (CASH) optimization problem
is defined as
\begin{align}
\!\!\!\!F^*\!\!,  \theta^*
=  \arg \min\nolimits_{ \theta  \in \Lambda_j, F_j \in \mathcal{F} } \!
\frac{1}{K} \sum\nolimits_{i = 1}^K
\min_{\vect{w}_j}
\mathcal{L}\left( F_j(\mathbf{w}_j; \theta_j), \mathcal{D}^i_{\text{train}}, \mathcal{D}^i_{\text{valid}} \right)
\!,\!
\label{eq:CASH_def}
\end{align}
where $\mathcal{L}\left( F_j(\mathbf{w}_j; \theta_j), \mathcal{D}^i_{\text{train}}, \mathcal{D}^i_{\text{valid}} \right)$
denotes the validation loss that $F_j$ achieves on 
$\mathcal{D}^i_{\text{valid}}$ with parameter $\vect{w}_j$, hyper-parameter $\theta_j$ and training data $\mathcal{D}^i_{\text{train}}$
\end{example}
\vspace{5px}

In this problem, the candidate models 
(e.g., classifiers in Scikit-Learn) and their corresponding hyper-parameters make up the search space in the CASH problem. However, the optimization of \eqref{eq:CASH_def} might be highly challenging. 
First,  the absence of an explicit expression for the objective impedes our understanding of its properties, such as the degree of smoothness, which is typically beneficial for traditional optimization problems. 
In addition, the decision variables, i.e., $\theta$ and $F_j$, 
may not be continuous, e.g., $\theta$ for kNN includes the number of nearest neighbors, which is a discrete variable. 
Finally, in order to estimate the 
validation loss, we need to train the model $F_j$ and update its parameter $\vect{w}_j$, 
which usually requires large computational cost and time consumption.
Next, we introduce four popular solutions with open-source code:
\begin{itemize}[leftmargin=*]
\item \textit{Auto-WEKA}\footnote{\url{https://www.cs.ubc.ca/labs/algorithms/Projects/autoweka/}} \cite{thornton2013auto} is designed for the automated selection and tuning of ML algorithms provided in the WEKA\footnote{\url{https://www.cs.waikato.ac.nz/ml/weka/}} library. 
Auto-Weka applies a Bayesian optimization method called  Sequential Model-based Algorithm Configuration (SMAC), which can search learning configurations for discrete variables. Besides, to balance the evaluation precison and computational cost, Auto-WEKA uses a reducing data quantity strategy (see Section \ref{sec:reduce_qd}). It discards underperforming parameter settings after evaluating them on a limited number of dataset folds.
\item \textit{Auto-Sklearn}\footnote{\url{https://www.cs.ubc.ca/labs/algorithms/Projects/autoweka/}} \cite{feurer2015efficient, feurer2022auto} is built on top of the popular scikit-learn library\footnote{\url{https://scikit-learn.org/}}. Auto-Sklearn 1.0 proposes k-nearest dataset (KND), a meta-learning approach  as the experienced technique to get better initialization and further improves performance by integrates ensemble methods. Compared to its 1.0 version, PoSH Auto-sklearn uses ``portfolios'' which is a meta-feature-free approach to warm start an AutoML system with less time consumption. Besides, it uses the successive halving bandit strategy to accelerate the evaluation process and obtain good indication of the generalization error. Furthermore, Auto-Sklearn 2.0 introduces an automated policy selection mechanism, which is designed to select model selection strategies like 
Cross-Validation Ensemble Selection \cite{caruana2006getting}, as well as budget allocation strategies (see Section \ref{sec:bandit}) including full budget and SuccessiveHalving \cite{karnin2013almost, jamieson2016non}, tailored for different datasets.
\item \textit{TPOT}\footnote{\url{https://automl.info/tpot/}}
 \cite{olson2016tpot} utilizes evolutionary algorithms (as mentioned in Section \ref{sec:evl_alg})
  to thoroughly investigate various potential pipelines and pinpoints the most effective one. In TPOT, machine learning operators are treated as primitives in evolutionary algorithm. These primitives are then assembled into functional machine learning pipelines using tree-based structures, and the evolutionary algorithm plays a crucial role in evolving these tree-based pipelines. This approach enables TPOT to explore a wide range of machine learning solutions effectively and efficiently.
\item \textit{H2O AutoML}\footnote{\url{https://h2o.ai/platform/h2o-automl/}} \cite{ledell2020h2o} is a fully automated supervised learning algorithm within the H2O framework\footnote{\url{https://github.com/h2oai}}. 
It combines the efficiency of random search with the robustness of stacked ensembles. 
This approach enables H2O AutoML not only matches but also exceeds the performance of other frameworks that use more complex model tuning techniques, 
such as Bayesian optimization or evolutionary algorithm.
\end{itemize}

We present the results of the AutoSklearn, TPOT and Random Search
on the part of datasets from previous AutoML challenges \cite{guyon2019analysis}. 
Random Search serves as the baseline, 
where 100 models and their hyperparameters were randomly selected from the open-source ML tool Data Science Machine (DSM) \cite{stadelmann2018deep}
for training, 
and the best result
is chosen as the final performance. 
As shown in Table \ref{tab:AutoML_results}, 
under similar time constraints, AutoML methods demonstrate a noticeable performance improvement compared to Random Search, 
especially on the Digits and Madeline datasets. This suggests that automatic machine learning can effectively address the CASH problem.

\useunder{\uline}{\ul}{}
\begin{table}[t]
\caption{Comparison of performance on multiple different tasks and datasets for Auto-sklearn, TPOT, and the random search methods (Adapted from~\cite{tuggener2019automated}), where ``Time'' is in minutes and ``Test'' denotes the performance on the testing dataset.}
	\vspace{-10px}
	\centering
	\footnotesize
	\resizebox{\columnwidth}{!}{%
	\begin{tabular}{lll|cc|cc|cc}
	\hline
	\multirow{2}{*}{Dataset} & \multirow{2}{*}{Task}    &\multirow{2}{*}{Metric}      & \multicolumn{2}{c|}{Random Search} & \multicolumn{2}{c|}{Auto-sklearn} & \multicolumn{2}{c}{TPOT} \\
							 &                 &               & Test             & Time            & Test                 & Time       & Test             & Time   \\ \hline
	Cadata                   & Regression     & Coefficient of determination            & 0.711           & 55.0            & {\ul 0.733}         & 54.9       & \textbf{0.799}  & 54.6  \\
	Digits                   & Multiclass  classification  & Balanced accuracy score    & 0.875           & 201.2           & \textbf{0.954}      & 201.2      & {\ul 0.948}     & 207.2 \\
	Fabert                   & Multiclass classification   &Accuracy score   & 0.867           & 77.5            & \textbf{0.891}      & 77.4       & {\ul 0.884}     & 78.5  \\
	Madeline                 & Binary classification    & Balanced accuracy score      & 0.768           & 48.3            & \textbf{0.890}      & 48.2       & {\ul 0.868}     & 53.0  \\
	Philippine               & Binary classification     & Balanced accuracy score     & 0.741           & 56.3            & {\ul 0.763}         & 56.2       & \textbf{0.770}  & 56.4  \\ \hline
	\end{tabular}%
	}
	\label{tab:AutoML_results}
\end{table}

\subsection{One-shot NAS}

One-shot NAS~\cite{cai2018proxylessnas, wu2019fbnet, zhang2020you} is a popular research area within the field of AutoML. 
This technique diverges from vanilla NAS methods~\cite{zoph2017learning, real2019regularized, luo2018neural} which typically involve training and evaluating numerous individual architectures separately. In one-shot NAS, all candidate architectures in the search space is often structured by a supernet which is trained once, and sub-networks (candidate architectures) are sampled from this supernet by inheriting corresponding weights. Therefore, this approach dramatically reduces the cost of architecture evaluation and
make NAS more accessible and practical for a wider range of applications. 

The supernet \cite{saxena2016convolutional,bender2018understanding, pham2018efficient} of one-shot NAS is a DAG (as mentioned in Section \ref{sssec:Directed acyclic graph}). 
Specifically, in the supernet, nodes denote latent representations (e.g., a feature map in CNN); edges correspond to operations in the network (e.g., convolution, pooling, recurrent units); each sub-network corresponds to a specific neural network architecture.
However, as the size of supernet grows, search algorithms like reinforcement learning or evolutionary algorithm cannot avoid requiring a large number of architecture evaluations \cite{liu2018darts}. To overcome this problem, the gradient-based method \cite{liu2018darts, chen2019progressive, chen2020stabilizing} is proposed for one-shot NAS. Instead of directly searching in a discrete search space, this method utilize search space transformation to make the search space continuous and optimizes the architecture by gradient descent of supernet on the validation set.

DARTS~\cite{liu2018darts} is a representative gradient-based one-shot NAS method. 
To make discrete choices in each edge differentiable, 
DARTS employs softmax transformation to transform the discrete selection of these choices into continuous learning on architecture hyperparameter $\alpha$ of supernet 
(see  \eqref{Eq:trans_darts} in Section \ref{sssec:transf_softmax}). 
After continuous relaxation, DARTS uses iterative differentiation (as mentioned in Section \ref{sec: grad method}) to jointly optimize the architecture parameters $\alpha$ (outer optimization) and supernet model weights $w$ (inner optimization) separately. 
 As shown in Algorithm \ref{alg:DARTS}, 
to avoid the expensive computation cost of solving the bi-level optimization, 
DARTS applies a single step of gradient descent to approximate the optimal model weights $w^*(\alpha)$ in inner optimization:
$w^*(\alpha) \approx w - \xi \nabla_w \mathcal{L}_\text{train} (w, \alpha)$.
Therefore, the one-shot NAS is achieved 
by  training this supernet model using
alternative gradient descent.
When the optimization is finished, DARTS determines the final model architecture by choosing the operation that corresponds to the maximum $\alpha$ value on every edge of supernet, and subsequently retrain the model from scratch.

\begin{algorithm}[t]
\caption{DARTS - Differentiable Architecture Search}
\footnotesize
\begin{algorithmic}
\label{alg:DARTS}
\STATE Initialize a supernet model $\mathcal{A}$ with weights $w$ and  hyperparameters $\alpha$.

\WHILE{not converged}
\STATE Outer optimization: update  $\alpha$ by descending $\nabla_\alpha \mathcal{L}_\text{val} (w^*(\alpha), \alpha), w^*(\alpha) \approx w - \xi \nabla_w \mathcal{L}_\text{train} (w, \alpha).$
\STATE Inner optimization: update  $w$ by descending $\nabla_w \mathcal{L}_\text{train} (w, \alpha)$.
\ENDWHILE
\STATE Obtain the final architecture: choose operations based on maximum $\alpha$ for each edge of $\mathcal{A}$.
\end{algorithmic}
\end{algorithm}

Actually, 
one-shot methods applying weight-sharing strategy is based on a fundamental assumption~\cite{liu2018darts}: 
the architecture rankings derived from the one-shot evaluation method consistently correlate with those obtained through their independent training. 
Subsequent works  against this assumption \cite{pourchot2020share, zela2019bench},
and reveal a coupling effect of $w$ and $\alpha$ within the one-shot framework \cite{guo2020single}, 
which causes problems in the one-shot model, including rank disorder and operational biases 
(e.g., tending to select skip connection operation~\cite{chu2020fair}).

\begin{figure}[t]
	\begin{minipage}{0.45\linewidth}
		\centerline{\includegraphics[width=0.93\textwidth]{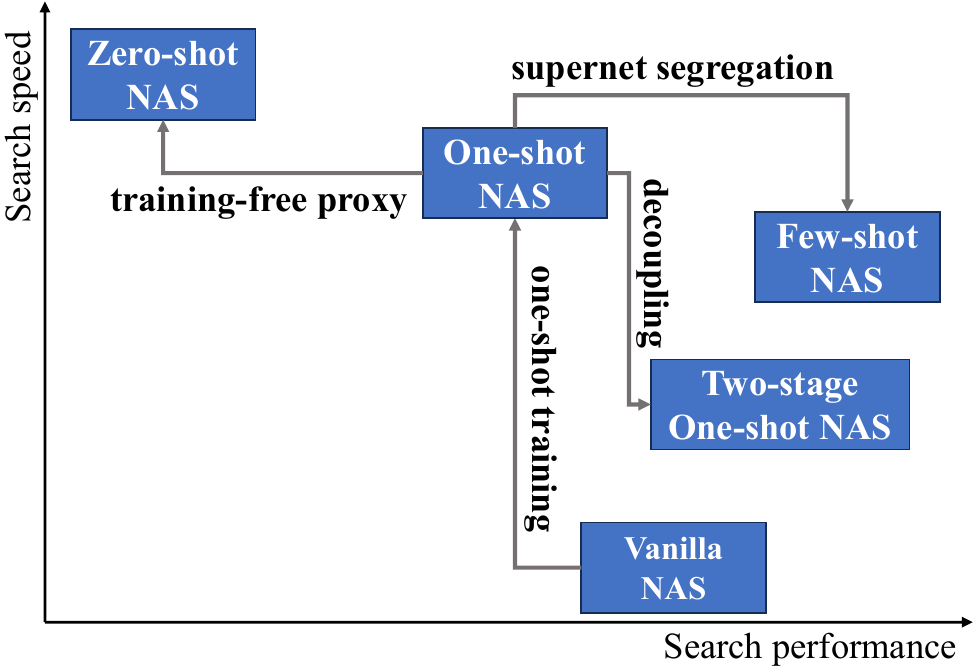}}
		\vspace{-3px}
		\caption{The technical roadmap of NAS.}
		\label{fig: NAS roadmap}
	\end{minipage}
	\makeatletter\def\@captype{table}\makeatother
	\begin{minipage}{0.54\linewidth}
		\vspace{-0.8cm}
		\centering
		\small
		\caption{Top: NAS results on CIFAR-10 using the NASNet search space (adapted from~\cite{zhao2021few}).
			Below: The performance of zero-shot and one-shot methods based on ProxylessNAS \cite{cai2018proxylessnas} search space on the ImageNet-1K dataset (adapted from~\cite{li2023zero}).}
		\label{tab: NAS results}
		\setlength\tabcolsep{2pt}
		\begin{tabular}{cccc}
			\hline
			Dataset&Method                                                                                              & Error       & GPU hours    \\ \hline
			\multirow{4}{*}{CIFAR-10} &\begin{tabular}[c]{@{}c@{}}NASNet-A~\cite{zoph2017learning}\\      \end{tabular}           & 2.65  & 2000          \\
			&\begin{tabular}[c]{@{}c@{}}DARTS~\cite{liu2018darts}\\      \end{tabular}   & 2.76±0.09          & 1          \\
			&\begin{tabular}[c]{@{}c@{}}Fair-DARTS~\cite{chu2020fair}\\      \end{tabular}         & 2.54±0.05           & 3 \\
			&\begin{tabular}[c]{@{}c@{}}few-shot DARTS~\cite{zhao2021few}\\      \end{tabular} & 2.31±0.08           & 1.35           \\ \hline
			ImageNet&\begin{tabular}[c]{@{}c@{}}Darts~\cite{liu2018darts}\\      \end{tabular}           & 74.39 & 200          \\
			-1K&\begin{tabular}[c]{@{}c@{}}Zen-score~\cite{lin2021zen}\\      \end{tabular}         & 71.78          & 1.6 \\ \hline
		\end{tabular}%
	\end{minipage}
\end{figure}

To alleviate the issue of coupling effect, the two-stage paradigm is proposed. 
Concretely, 
In the the two-stage paradigm method \cite{bender2018understanding}, the one-shot NAS process involves training the complete supernet in the first stage, and in the second stage, 
it evaluates specific architectures by zeroing out large portions of the one-shot model with a designed dropout strategy. 
However, this method only addresses the coupling between $w$ and $\alpha$ but the
coupling problem in $w$ still remains. 
To address this, SPOS \cite{guo2020single} trains only one single path  uniformly sampled from the supernet in the first stage.  Fair-DARTS~\cite{chu2020fair} is proposed 
as a collaborative competition approach which ensures each operation has an equal opportunity to develop its strength. 
Since the two-stage methods inherit the parameters of the supernet and only perform inference in the second stage, the search process remains relatively efficient.
Instead of using a two-stage process, Few-shot NAS \cite{zhao2021few} divides the supernet into few-shot sub-supernets to reduce the coupling effect. During the architecture search phase, this method efficiently chooses the most promising sub-supernet as a starting point to determine the final architecture. Besides, recent works~\cite{zhang2022gradsign, mellor2021neural} have proposed zero-shot NAS, a method that ranks candidate architectures in the initial phase without training the models. Although, zero-shot NAS, such as Zen-score~\cite{lin2021zen} and Gradnorm~\cite{abdelfattah2020zero}, considerably cuts down time cost, the low accuracy performance cannot be overlooked \cite{li2023zero}. The technical roadmap of NAS is shown in Figure \ref{fig: NAS roadmap}, which reveals the trade-offs between search performance and search speed for different NAS methods. 

Tables~\ref{tab: NAS results} present the experimental results of the NAS methods. 
As indicated in the results, DARTS can greatly reduce the time required for vanilla NAS. 
Meanwhile, few-shot DARTS makes a better trade-off between the vanilla NAS and one-shot NAS. 
It achieves the lowest error while its time cost is closest to that of DARTS. 
Besides, the experimental results reveal that although zero-shot NAS considerably cuts down time cost, the performance of searched architecture is not excellent enough. Therefore, a practical approach for real-world applications might involve using zero-shot methods for an initial, broad screening of the search space, followed by more detailed searches with few-shot or two-stage methods for precision.

\subsection{AutoML for Foundation Models}

The term ``foundation model'' is defined as ``any model that is trained on broad data (generally using self-supervision at scale) that can be adapted 
(e.g., fine-tuned) to a wide range of downstream tasks'' by the Stanford Institute for Human-Centered Artificial Intelligence's (HAI) Center \cite{bommasani2022opportunities}. 
At present, representative foundation models include the large language model (LLM) ``GPT-x'' series \cite{brown2020language}, the text-to-image generation model ``Stable Diffusion'' \cite{rombach2022high} and the graph learning model ``GROVER'' \cite{rong2020self}.

\begin{figure}[t]
	\centering
	\includegraphics[width=0.80\textwidth]{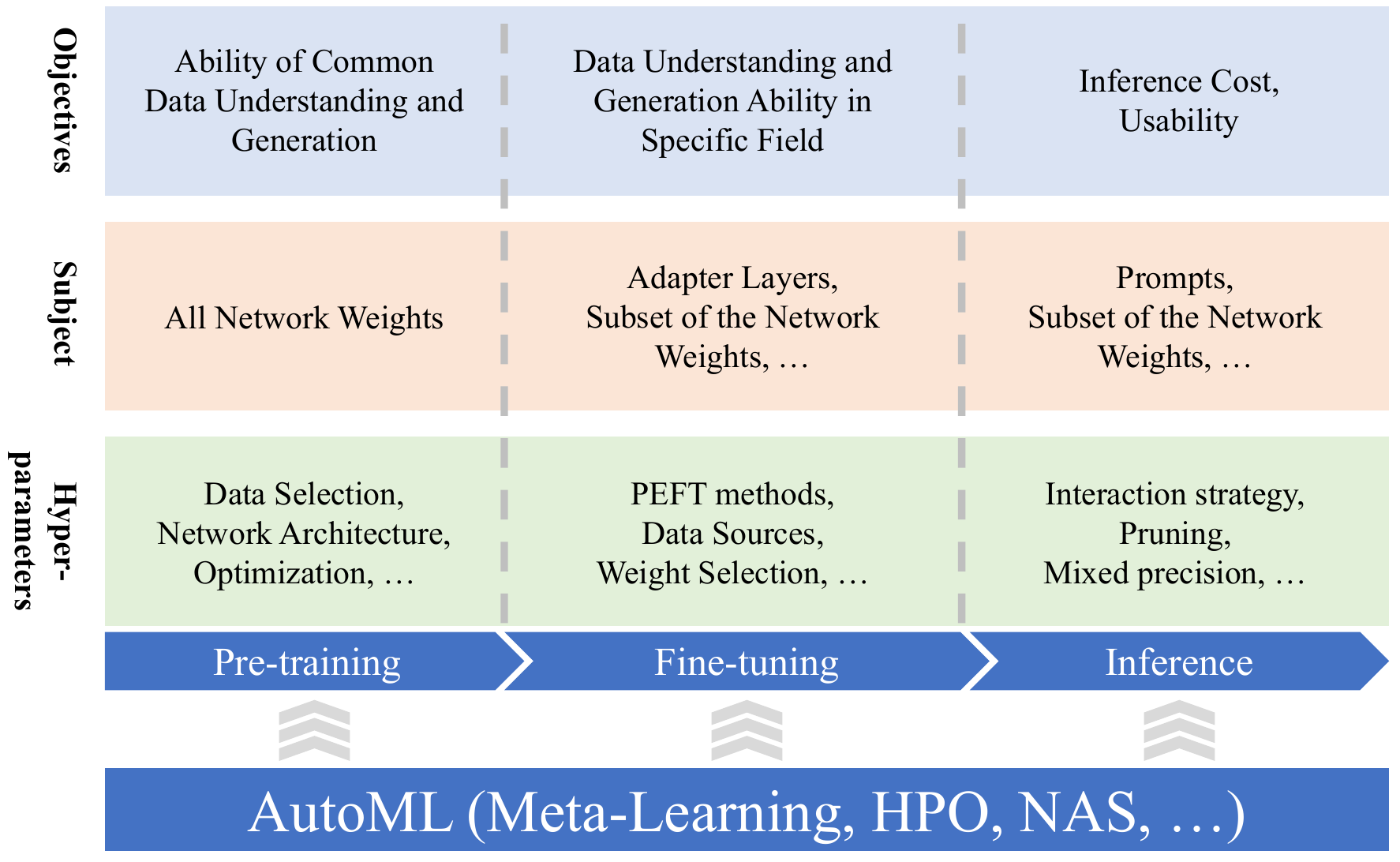}
	\caption{AutoML can be used in all stages of foundation model lifecycle. In each stage, AutoML optimzes both subject and associated hyper-parameters to achieve corresponding objective \cite{tornede2023automl}.}
	\label{fig:fdm_automl}
\end{figure}

The lifecycle of a foundation model involves three stages: pre-training, fine-tuning, and inference~\cite{zhou2023comprehensive, liu2023towards}. The development of foundation models in their lifecycle is considerably more complex than task-specific models, but the process still primarily relies on manual labor and expert knowledge \cite{tornede2023automl}. However, it is certain that AutoML can play a role throughout the entire lifecycle of foundation models (as shown in Figure \ref{fig:fdm_automl}).
Without loss of generality, we take the LLM as an example to briefly introduce the three stages in the lifecycle of foundation models and the role AutoML can play in each of these stages.

\begin{itemize}[leftmargin=*]
	\item \textbf{Pre-training stage}: the LLM is trained on a vast corpus of unlabeled text with self-supervised learning to learn language patterns and general knowledge.   
	At the begin of this stage, the learning configurations of LLM need to be determined. These learning configurations encompass various training hyperparameters, such as learning rate schedule \cite{loshchilov2016sgdr, gotmare2018closer}, optimizer \cite{duchi2011adaptive, kingma2015adam}, and batch size, as well as aspects of the model's architecture, including types of attention mechanism \cite{vaswani2017attention} , the number of attention heads, embedding dimensions, activation functions \cite{nair2010rectified}, and the selection and combination of self-supervised algorithms \cite{kenton2019bert}. Consider the pre-training stage is typically the most resource-intensive phase and the learning configurations have a crucial impact on pre-training performance, 
	efficiently finding a set of optimal learning configurations is important. 
	AutoML is one possible way to achieve this goal, for example, we can use evolutionary algorithm \cite{huetal2021ranknas} to search more powerful LLM architectures.
	\item \textbf{Fine-tuning stage}: after pre-training, the LLM is further trained on specific data related to certain domains or tasks through supervised learning. This process fine-tunes the LLM's knowledge, making it more specialized and adaptable to specific use cases. Parameter-Efficient Fine-Tuning (PEFT) methods \cite{li2021prefix, hu2022lora} enable rapid adaptation of models to downstream tasks and scenarios by optimizing a small set of parameters. The selection of PEFT methods and parameters for updating significantly influences the  fine-tuning effectiveness. In \cite{ding2023parameter}, four representative PEFT methods — Prompt-tuning \cite{liu2022p}, Prefix-tuning \cite{li2021prefix}, LoRA \cite{hu2022lora}, and Adapter \cite{houlsby2019parameter} — are evaluated across over one hundred natural language processing tasks. The experimental results reveal that no single PEFT method consistently outperformed across all tasks. Moreover, combining multiple PEFT methods could further enhance performance. Additionally, the choice of parameters for updating, both in proportion and in the number of layers, affects fine-tuning performance. Therefore, a reasonable idea is to apply AutoML to search suitable PEFT methods \cite{wang2023search, zhou2023autopeft} and the parameters \cite{liu2021autofreeze} that require updating for corresponding downstream tasks.
	\item \textbf{Inference stage}: the fine-tuned LLM is used for generating text in downstream language-related tasks like question answering and information extraction. 
	However, given the substantial parameter count and high computational cost of foundation models, a critical challenge is to make their inference more cost-effective and efficient. From a structural standpoint, NAS can be utilized to compress a well-trained foundation model \cite{chen2021adabert, kwon2022fast}. From an interactive perspective, hyperparameter setting is necessary during the inference stage. Hyperparameters, such as ``temperature'' for controlling randomness and ``max\_token'' for determining generation length, significantly influence the model's output \cite{wang2023cost}. Therefore, employing AutoML to generate optimal hyperparameters for different tasks and foundation models can enhance inference performance while reducing operational costs.
\end{itemize}

In fact, there are already studies applying AutoML techniques to optimize each stage in the lifecycle of foundational models \cite{zhou2023autopeft, wang2023cost}. Here, we introduce five works that use AutoML to obtain better fine-tuning strategy in the fine-tuning stage:

Existing methods \cite{houlsby2019parameter, li2021prefix, xia2022towards, hu2022lora} often employ a fixed fine-tuning strategy, overlooking task-specific adaptation. To address this issue, Hyperformer++ \cite{karimi2021parameterefficient} trains a hypernetwork through multi-task and multi-layer weight sharing strategy, which helps the hypernetwork learn how to generate task- and layer-specific  parameters of Adapter \cite{houlsby2019parameter} for transformer based on the input task and the number of layers id embedding.
Compared to Hyperformer++ which only considers a single PEFT method, the search space of AutoPEFT \cite{zhou2023autopeft} encompasses multiple PEFT methods, such as serial adapters, parallel adapters, and prefix-tuning. By employing multi-dimensional Bayesian optimization, AutoPEFT can identify the optimal PEFT configuration that balances cost and performance in a single run (as shown in Table \ref{tab: AutoPEFT}).
Besides, S2PGNN \cite{wang2023search} considers the key modules of GNN fine-tuning as its search space and employs gradient-based search technology to identify the optimal fine-tuning strategy for pre-trained GNN. 
In addition to selecting modules for fine-tuning, determining the appropriate timing for fine-tuning or freezing model parameters can reduce both the time spent on fine-tuning and the risk of overfitting. AutoFreeze \cite{liu2021autofreeze} addresses this by developing a gradient-norm-based test that ranks layers according to their changes in SVCCA \cite{Raghu2017svcca}. Building on this, AutoFreeze then selects the layers with the slowest rate of change for freezing.
Moreover, Quicktune \cite{arango2024quicktune} focuses on identifying the most suitable pre-trained model and its optimal fine-tuning parameters for a new dataset within a limited time cost. Quicktune employs the Gray-Box Bayesian Optimization method for both model selection and hyperparameter search and utilizes meta-learning to facilitate rapid transfer across tasks.

\begin{table}[tbh]
	\caption{Results on the GLUE benchmark with BERT as backbone. Adamix \cite{wang2022adamix} denotes a mixture strategy of adaptation modules. AutoPEFT achieved better results than full parameter optimization (FFT) and Adamix, with slightly larger fine-tuning parameter using than the PEFT method  (adapted from~\cite{zhou2023autopeft}).}
	\label{tab: AutoPEFT}
	\footnotesize
	\begin{tabular}{lccccc}
	\hline
	Method   & \# Param. & RTE        & MRPC       & STS-B      & CoLA       \\ \hline
	FFT      & 100\%     & 71.12±1.46 & 85.74±1.75 & 89.00±0.45 & 59.32±0.62 \\
	Prefix   & 0.17\%    & 70.54±0.49 & 85.93±0.89 & 88.76±0.15 & 58.88±1.15 \\
	LoRA     & 0.27\%    & 65.85±1.49 & 84.46±1.04 & 88.73±0.08 & 57.58±0.78 \\
	Adamix   & 0.81\%    & 70.11±0.62 & 86.86±1.12 & 89.12±0.11 & 59.11±1.00 \\ \hline
	AutoPEFT & 1.40\%    & 72.35±0.94 & 87.45±0.87 & 89.17±0.24 & 60.92±1.47 \\ \hline
	\end{tabular}
	\end{table}

\section{Emerging Directions}
\label{sec:future}

In this section, we will talk about emerging directions of AutoML in four research focuses: problem setup, techniques, theory, and applications.

%
%
%
%
%
%
%

\subsection{Problem setup}
Current AutoML methods are primarily designed for 
classical machine learning problems. 
However, 
emerging learning problems in practical applications sometimes exceed the scope of the classical machine learning problem, e.g., the lack of labeled training data.
Compared to the classical machine learning problem, 
problem settings of AutoML is different for these emerging learning problems. 
For instance, few-shot learning \cite{wang2020generalizing} refers to the challenge of training machine learning models that can generalize to new tasks with limited labeled data. 
Therefore, the AutoML framework focuses on the automated identification or formulation of optimal meta-learning algorithms \cite{elsken2020meta}.  
In the realm of positive-unlabelled learning \cite{bekker2020learning}, the challenge lies in training models to differentiate between positive and negative classes, relying solely on positively labeled and unlabeled data. 
This requires the AutoML framework to automate the selection or creation of suitable methods for positive prior estimation and label propagation algorithms \cite{saunders2024automated}. 
Furthermore, for foundation models, the AutoML problem setup entails the automatic selection or construction of appropriate pre-training self-learning tasks, as well as the optimization of hyperparameters and fine-tuning strategies \cite{arango2024quicktune}. 
Although recent advancements have seen the application of AutoML to these emerging learning problems, further research and exploration are still needed.

\subsection{Techniques}
Efficiently finding a set of learning configurations that is as good as possible from a complex search space is the main challenge of AutoML. Several emerging technologies are proposed to tackle this challenge.
For example, some researchers have explored to reduce the complexity of search space. Specifically, these works reduce the size of search space by using meta-learning to transfer knowledge from previous tasks \cite{elsken2020meta,zhao2021few}, and decomposing the search space into smaller and more manageable subspaces \cite{li2023volcanoml}. 
Besides, researchers have proposed methods to improve the evaluation efficiency and robustness. 
For example, Landmark regularization is proposed by \cite{yu2021landmark} to improve the consistency between the performance rankings of the supernet model and that of the standalone architectures.
$\xi$-GSNR \cite{sun2023unleashing}  uses gradient information to predict the ranking of candidate architectures 
without training them.
MixPath \cite{chu2023mixpath} proposed a shadow batch normalization  to improve the robust of architecture evaluation in multi-path one-shot NAS.
Although progress has been made, these emerging techniques also pose new challenges and open questions that require further investigation and experimentation. Specifically, the reduction in search space may sometimes lead to overlooking potential models, 
and the accuracy of performance predictors sometimes hinges on the quality and representativeness of the data they are trained on.

\subsection{Theory}
As described in Section \ref{sec:overview}, the AutoML task, such as NAS or ML pipeline configuration, can be modeled as a bi-level optimization problem. Therefore, theoretical research on bi-level optimization can often bring new breakthroughs to the field of AutoML. 
In the optimization theory field, convergence speed is a critical problem for AutoML and matters a lot in the choice of techniques for the optimizer, as the evaluation of one learning configuration usually requires a model training, which is very expensive. 
While many theories of convergence have been developed for basic techniques,
e.g., derivative free optimization~\cite{conn2009introduction,hashimoto2018derivative} and 
gradient descent~\cite{bengio2000gradient, sow2022convergence}, 
it is still not clear how fast they can identify a good configuration.
In the field of learning theory, it is important to clarify the generalization ability for a specific AutoML approach. 
In section~\ref{sec:theory}, we provide qualitative generalization performance analysis of AutoML methods through error decomposition. 
Currently, there are also some works provide generalization performance analysis for specific AutoML methods~\cite{cortes2017adanet, couellan2018bi}. 
However, more comprehensive and quantitative generalization analysis is needed to understand the generalization ability of AutoML methods.

\subsection{Applications}

In recent years, there has been a notable surge in the adoption of AI to assist scientific research across various domains \cite{vamathevan2019applications, jones2017machine, karagiorgi2022machine}. Within this context, AutoML can further unleash the potential of AI techniques, helping scientists expedite research, enhance precision, and uncover insights that are beyond human analytical capacity. 
Specifically, in biomedical sphere, AutoML is revolutionizing drug discovery and genomics, expediting molecular activity prediction, optimizing drug formulations, and enabling personalized treatment plans \cite{schneider2018automating}. 
Likewise, within the realm of oncology predictive modeling, an AutoML-based tool, JADBio \cite{tsamardinos2022just}, has been introduced to generating predictive models with high-performance criteria.
In the field of edge computing \cite{iyer2023ai}, more lightweight ML models are needed for fast and accurate inference. 
AMC \cite{he2018amc} utilizes AutoML techniques to replace manual efforts in achieving automatic compression of models with high compression rate.
In the disciplines of earth and environmental sciences, AutoML has demonstrated its efficacy in characterizing unconventional reservoirs, accurately predicting key properties like shale volume, porosity, and bitumen content \cite{mubarak2023hierarchical}. 
Moreover, in the field of materials science, AutoML is applied in predicting the properties of concrete, metals, and fiber-reinforced polymers, facilitating the discovery and optimization of novel materials \cite{conrad2022benchmarking}. It can be foreseen that with the further popularization and deepening of AI in scientific research, ``AI for Science'' will gradually become a promising application direction of AutoML.

\section{Conclusion}


Automated machine learning (AutoML) targets at substituting what human do in the machine learning process. 
Its primary objective is the automated determination of optimal learning configurations tailored to a given machine learning task.
In this survey, we provide a systematic review of AutoML, from principles to practices. 
We first generate the formal definition of AutoML from that of machine learning.
Then we formulate the AutoML problem as a bi-level optimization problem, propose the general learning strategy, and provide theoretical interpretation of AutoML. 
Based on above principles, we present an exposition on AutoML practices. 
We categorize existing AutoML approaches based on three main factors of AutoML~(search space, search algorithm and evaluation strategy) and further introduce representative approaches under each category. 
Finally, we make a discussion on exemplary applications and the emerging directions of AutoML. 

\section*{Acknowledgments}
The first four
authors are ordered accordingly to their contributions. 
We thank the assistance of Hongyi Nie for writing this survey. 

\bibliographystyle{ACM-Reference-Format}
\bibliography{sample-base}

\end{document}